\documentclass[lettersize,journal]{IEEEtran}
\usepackage{amsmath,amsfonts}
\usepackage{array}
\usepackage[caption=false,font=normalsize,labelfont=sf,textfont=sf]{subfig}
\usepackage{textcomp}
\usepackage{stfloats}
\usepackage{url}
\usepackage{verbatim}
\usepackage{graphicx}
\usepackage{cite}
\usepackage{array}
\usepackage{color}
\usepackage{textcomp}
\usepackage{stfloats}
\usepackage{url}
\usepackage{verbatim}
\usepackage{graphicx}
\usepackage{cite}
\usepackage{mathrsfs}
\usepackage{euscript}
\usepackage[colorlinks, citecolor=black]{hyperref}
\usepackage{mathrsfs}
\usepackage{multirow,diagbox}
\usepackage{verbatim}
\usepackage{color}
\usepackage{colortbl}
\usepackage{makecell}
\usepackage{cellspace}
\usepackage{stfloats}
\usepackage{longtable}
\usepackage{booktabs}
\usepackage{array,multirow,multicol,graphicx}
\usepackage{color}
\usepackage{float}
\usepackage{multirow}
\usepackage[font=footnotesize]{caption}
\usepackage{subcaption}
\usepackage{lipsum}
\usepackage{array, tabularx, boldline}
\usepackage{eurosym}
\usepackage{amstext} 
\usepackage{amssymb} 
\usepackage{tabularray} 
\usepackage{adjustbox}
\usepackage[ruled,vlined,linesnumbered]{algorithm2e}

\definecolor{mygray}{gray}{0.6}
\definecolor{myblue}{rgb}{0.8,0.85,1}
\usepackage{array}
\usepackage{ragged2e}
\newcolumntype{L}[1]{>{\raggedright\let\newline\\\arrayback\textbf{SLA}sh\hspace{0pt}}m{#1}}
\newcolumntype{C}[1]{>{\centering\let\newline\\\arrayback\textbf{SLA}sh\hspace{0pt}}m{#1}}
\newcolumntype{R}[1]{>{\raggedleft\let\newline\\\arrayback\textbf{SLA}sh\hspace{0pt}}m{#1}}

\newcommand{\reffig}[1]{Figure \ref{#1}}
\newcommand{\reftable}[1]{Table \ref{#1}}

\definecolor{mygray}{gray}{0.9}

\usepackage{tikz}
\usepackage{amsmath}
\usepackage{filecontents}
\usepackage{threeparttable}
\usepackage[utf8]{inputenc} 
\usepackage[T1]{fontenc}    
\usepackage{url}            
\usepackage{booktabs}       
\usepackage{amsfonts}       
\usepackage{nicefrac}       
\usepackage{microtype}      
\usepackage{xcolor}         
\usepackage{graphicx}
\usepackage{amssymb}
\usepackage{times}
\usepackage{epsfig}
\usepackage{enumitem}
\usepackage{multirow}
\usepackage{makecell}
\usepackage{dsfont}

\DeclareMathOperator{\logotransfer}{\texttt{logo\_transfer}}
\DeclareMathOperator{\calmask}{\texttt{cal\_mask}}
\DeclareMathOperator{\pad}{\texttt{pad}}

\DeclareMathOperator{\randominit}{\texttt{random\_init}}
\DeclareMathOperator{\randomselect}{\texttt{random\_select}}
\DeclareMathOperator{\alphaschedule}{\texttt{alpha\_schedule}}
\DeclareMathOperator{\StyleReferenceSelection}{\texttt{style\_reference\_selection}}
\DeclareMathOperator{\logostyletransfer}{\texttt{logo\_style\_transfer}}

\DeclareMathOperator{\append}{\texttt{append}}

\allowdisplaybreaks[3]

\usepackage{algpseudocode}



\usepackage{siunitx}
\def\eg{\emph{e.g.,}\xspace}

\def\ie{\emph{i.e.,}\xspace}
\def\etal{\emph{et al.}\xspace}

\begin{document}

\title{Query-Efficient Video Adversarial Attack with Stylized Logo on Service Computing}

\author{Duoxun Tang\IEEEauthorrefmark{2}, Yuxin Cao\IEEEauthorrefmark{2}, Xi Xiao\IEEEauthorrefmark{1}, Derui Wang, Sheng Wen and Tianqing Zhu

\thanks{
\IEEEauthorrefmark{2} Duoxun Tang and Yuxin Cao contributed equally to this paper.

\IEEEauthorrefmark{1}Xi Xiao is the corresponding author.

Duoxun Tang and Xi Xiao are with Shenzhen International Graduate School, Tsinghua University, China. Emails: tdx25@mails.tsinghua.edu.cn, xiaox@sz.tsinghua.edu.cn.
Yuxin Cao is with the School of Computing, National University of Singapore, Singapore. Email: yuxincao@u.nus.edu. 
Derui Wang is with the Cybersecurity and Quantum Systems Group, CSIRO's Data61, Australia. Email: derek.wang@data61.csiro.au.
Sheng Wen is with the School of Science, Computing and Engineering Technologies, Swinburne University of Technology, Melbourne, VIC, Australia. Email: swen@swin.edu.au. 
Tianqing Zhu is with the Faculty of Data Science, City University of Macau. Email: tqzhu@cityu.edu.mo.
}

}

\IEEEtitleabstractindextext{
\begin{abstract}
\justifying\let\raggedright\justifying
In service computing, video classification has become a fundamental component of many intelligent applications. While Deep Neural Networks (DNNs) have demonstrated excellent performance in accurately recognizing video content, recent studies have shown that DNNs are highly vulnerable to adversarial examples. Therefore, a deep understanding of adversarial attacks can better respond to emergency situations. In order to improve attack performance, many style-transfer-based attacks and patch-based attacks have been proposed. However, the global perturbation of the former will bring unnatural global colors, while the latter is difficult to achieve success in targeted attacks due to the limited perturbation space. Moreover, compared to a plethora of methods targeting image classifiers, video adversarial attacks remain relatively underexplored. Therefore, to generate adversarial examples with a low budget and to provide them with a higher verisimilitude, we propose a novel black-box video attack framework, called Stylized Logo Attack (\textbf{SLA}). \textbf{SLA} is conducted through three stages. The first stage involves building a style reference set for logos, which can not only make the generated examples more natural, but also carry more target class features in targeted attacks. Then, Reinforcement Learning (RL) is employed to determine the style reference and position parameters of the logo within the video, which ensures that the stylized logo is placed in the video with optimal attributes. Finally, perturbations are optimized in a step-by-step manner so as to improve the fooling rate. Sufficient experimental results indicate that \textbf{SLA} can achieve better performance than state-of-the-art methods and still maintain good deception effects when facing various defense methods. We believe that the proposed \textbf{SLA} can raise awareness among the security community about the reliability and security of video classification systems and serve as a memorandum of possible attack methods for the future.
\end{abstract}

}

\maketitle

\IEEEdisplaynontitleabstractindextext

\section{INTRODUCTION}
In service computing, video understanding tasks using DNN-based models can be seen everywhere, such as video action recognition \cite{ji20123d}, video object detection \cite{zhao2019object}, and video segmentation \cite{nilsson2018semantic}. These models have been widely applied in the fields of service computing such as autonomous driving \cite{kuutti2020survey}, industrial vision \cite{prunella2023deep}, or medical science \cite{zhou2021review}. However, a long history of studies has indicated that DNNs are highly susceptible to adversarial attacks \cite{goodfellow2014explaining, wei2019sparse, szegedy2014intriguing}, posing security issues when applied in real-world services. Furthermore, as one of the most prevalent media for information dissemination in today's society, video facilitates the rapid spread of information but also introduces security risks associated with harmful content, including violence, pornography, hatred, terrorism, and misinformation~\cite{cao2026failures}. Such content can threaten the physical and mental health of minors and lead to politically sensitive controversies, public panic, or other adverse consequences. Moreover, attackers can maliciously manipulate pixels in illegal videos to amplify these risks. Thus, equal attention must be given to attack methods in order to better understand adversarial attacks and prepare for potential mitigations. In recent years, designing attack methods from the perspective of unrestricted perturbations has become an active area of research \cite{yan2021efficient, jia2020adv, chen2022attacking, yang2020patchattack, croce2022sparse, cao2023stylefool, gong2023palette, gong2023kaleidoscope,cao2024localstylefool}, with the aim of drawing the attention of the security community and simulating potential future attacks.

From the external manifestations of attacks, existing unrestricted attacks can be roughly divided into two categories: style-transfer-based attacks~\cite{duan2020adversarial, cao2023stylefool} and patch-based attacks~\cite{yang2020patchattack, croce2022sparse}. Specifically, both attack types involve superimposing a finely designed perturbation (global or local) on a clean sample to induce the classifier to make incorrect judgments. In extreme cases, modifying only one pixel of a clean sample can achieve the attack~\cite{su2019one}. For style-transfer-based attacks, their capability to alter all pixels can render the overall style of the sample unnatural. Compared to style-transfer-based attacks, patch-based attacks introduce perceptible but spatially localized changes, which are more practicable in the physical world. However, perturbation constraints such as $\ell_p$-norms are unnecessary for patch-based attacks since the perturbed region is already limited. Because patch-based attacks modify only small sub-regions, a key limitation is their relatively poor performance in targeted attacks~\cite{jia2020adv, yan2021efficient, chen2022attacking}. Therefore, attackers must devise strategies to move the examples across the decision boundary using regional patches.

Some white-box attacks \cite{inkawhich2018adversarial, li2019stealthy, chen2021appending, chang2022RoVISQ} have achieved remarkable success due to their full access to the model information. However, real-world attackers often have no access to the structure and parameters of the target model and instead operate under stricter settings, such as label-only \cite{brendel2018decisionbased,Cheng2018QueryEfficientHB,chen2020hopskipjumpattack,chen2020boosting,chen2020rays} or score-based \cite{ilyas2018black, liu2019signsgd, IEM2018PriorCB, guo2019simple, andriushchenko2020square} assumptions. One significant drawback of many current black-box attacks compared to white-box attacks is that the former often require a large number of queries to find adversarial examples. Additionally, this issue is exacerbated when conducting targeted attacks. Therefore, considering the weak targeted attack effectiveness of patch-based attacks and the query-unfriendly nature of black-box settings, designing an adversarial patch that can efficiently launch targeted attacks in a black-box setting presents a major challenge for attackers.

To address the challenges mentioned above, we propose a brand new attack framework called Stylized Logo Attack (\textbf{SLA}) against video classification systems based on a stylized patch. Operating under a black-box setting, \textbf{SLA} combines the strengths of style-transfer-based and patch-based attacks, forming a style-transfer-based patch attack that is efficient in queries and demonstrates satisfactory fooling rates in both targeted and untargeted attacks. 
Unlike existing stylized attacks that perturb entire videos globally, \textbf{SLA} perturbs only sub-regions of the video and stylizes them to make the adversarial video approach or even surpass the decision boundary of the classifier.
Specifically, our attack consists of three stages. Firstly, we construct a style set using a refined random search strategy, and introduce a logo set as candidate patch patterns. 
Secondly, we employ reinforcement learning (RL), which enables an agent to learn an optimal policy through interaction with an environment, to search for optimal logo attributes (\eg pattern, style, and position) and to overlay the resulting stylized logo onto appropriate regions in the clean video to produce adversarial examples. The search is guided by a reward function that considers classifier confidence, logo size, and coordinates to maximize attack effectiveness while preserving visual naturalness.
Lastly, we include a perturbation optimization stage based on random search to mitigate the weakness of existing patch-based and RL-based methods in search space inadequacy. In extensive comparative experiments conducted on four mainstream datasets (UCF-101 \cite{soomro2012ucf101}, HMDB-51 \cite{kuehne2011hmdb}, Kinetics-400 \cite{kay2017kinetics} and Kinetics-700 \cite{smaira2020short}) against existing patch-based attacks, \textbf{SLA} demonstrates excellent performance in both targeted and untargeted attacks, while retaining semantic integrity. Moreover, we evaluate the robustness of \textbf{SLA} against three defenses. 

This paper builds upon our earlier conference paper \cite{cao2024logostylefool}. The new contributions it presents are outlined below:

\begin{itemize}
    \item We introduce an innovative stylized patch-based attack framework, termed \textbf{SLA}, which employs a random search strategy for perturbation updates. This framework accomplishes attacks by superimposing extra stylistic attributes onto patches within clean videos. It allows perturbations to closely approach decision boundaries through stylization, while remaining imperceptible in localized sub-regions. 
    \item We utilize RL to search for the optimal way to overlay patches, defining a new search space for patch attributes. Correspondingly, we design a specific reward function that considers the impact of patch size and position to enhance naturalness. 
    \item We supplement a perturbation optimization stage that utilizes square-shaped random search after RL to overcome the limited effectiveness of patch-based attacks in targeted scenarios and the restricted search space of RL, which can ensure the feasibility of targeted attacks in \textbf{SLA}.
    \item Besides UCF-101 and HMDB-51, we extend \textbf{SLA} to two larger datasets, Kinetics-400 and Kinetics-700.
    Our approach outperforms state-of-the-art methods in representative metrics. Additionally, it exhibits optimal robustness when faced with different defensive methods.
    \item This paper presents more comprehensive attack examples across various methods in different scenarios, and adds two sets of user studies to verify imperceptibility of the adversarial videos generated by \textbf{SLA}.
\end{itemize}

\section{RELATED WORK}
\label{sec:RELATED WORK}
\subsection{Restricted Attacks}
To ensure the imperceptibility of perturbations, many studies \cite{jiang2019black,wei2020heuristic,kumar2020finding,pony2021over,jiang2023towards} generate adversarial videos $x_{adv}$ from clean videos $x$ to fool the DNN-based models, while keeping $x_{adv}$ within the $\varepsilon$-ball of $x$ ($\|x_{adv}-x\|_p \le \varepsilon$). Some white-box attacks \cite{wei2019sparse,xu2022sparse,wu2023imperceptible} assume that the attacker has full access to the specific structure of the victim models, allowing them to get precise information such as gradients. However, motivated by real-world scenarios, the black-box setting, where the attacker has no access to the model parameters or gradients, has become the main research topic. Following this setting, Jiang \etal \cite{jiang2019black} proposed the pioneering video attack method, V-BAD, which utilizes tentative perturbations and partition-based rectifications. To overcome the expensive computation of dense attacks and further enhance the invisibility of perturbations, Wei \etal \cite{wei2020heuristic} proposed a sparse attack for video classification models, which heuristically perturbs a subset of frames. With the development of RL, some attacks \cite{wang2021reinforcement, wei2022sparse, wei2023efficient, chen2023coreset} incorporate RL to search for key regions or key frames of the video.
However, these methods suffer from high query costs or low-efficiency in targeted attacks, which makes them far from being applicable in real-world scenarios.

\subsection{Style-transfer-based Attacks}
Style transfer techniques \cite{hertzmann2023image, gatys2015neural} facilitate the development of style-transfer-based attacks, which first appeared in the image domain, with AdvCam \cite{duan2020adversarial} being a representative example. Under constraints such as content loss, style loss \cite{ruder2018artistic} and smoothness loss \cite{johnson2016perceptual}, adversarial examples are crafted to resemble legitimate natural styles. However, for videos, style-transfer-based attacks typically need to balance the content and style of the stylized video while considering temporal consistency between frames \cite{huang2017real}.

StyleFool \cite{cao2023stylefool} was the first to utilize style-transfer-based attacks for videos without perturbation restrictions, incorporating temporal loss in addition to the classic constraints mentioned above. It also introduced style references that carry target class information to approach the decision boundary of the target class, achieving notable success in targeted attacks. However, a common drawback of style-transfer-based attacks is that global perturbations may inevitably result in unrealistic object colors and decrease the naturalness of adversarial videos.

\subsection{Patch-based Attacks}
Patch-based attacks \cite{yang2020patchattack,jia2020adv,chen2022attacking,croce2022sparse} typically add perceptible patches to local regions of clean examples, effectively misleading classifiers while avoiding the global artifacts commonly introduced by style transfer. Early naive methods such as HPA \cite{fawzi2016measuring} used solid-color patches as perturbations, but proved ineffective for targeted attacks. Subsequent research indicated that convolutional neural networks (CNNs) are particularly sensitive to striped perturbations \cite{yin2019fourier}. Square-Attack (SA) \cite{andriushchenko2020square} leveraged this insight by initializing inputs with vertical stripes and then conducting a random search for square perturbations.

The attributes of patches can be structured into a search space, leading many patch-based attacks to employ RL for optimization. Specifically, given a search space, RL utilizes an agent to execute policies and adjust them based on environmental feedback. The attributes (\eg position, shape, transparency, size) of the adversarial patch constitute the search space, turning the attack into a decision-making process for the agent. RL-based methods have become increasingly important in adversarial machine learning since RL reframes perturbation generation from random search into a structured optimization problem, thus enhancing attack efficiency. Several representative attacks have been proposed. For example, PatchAttack \cite{yang2020patchattack} targets image classifiers and achieves over 99\% fooling rate in untargeted attacks on ImageNet \cite{russakovsky2015imagenet}, but only about 10\% fooling rate in targeted attacks. Moreover, many real-world patterns (\eg bullet comments, watermarks, logos) carry natural semantics, enabling patch-based attacks to exploit these cues for greater visual plausibility. Adv-watermark \cite{jia2020adv} and BSC \cite{chen2022attacking} both utilize semantic patches for attack purposes. However, the transparency of the Adv-watermark sometimes reduces its attack effectiveness. While BSC achieves natural and inconspicuous effects by adding common bullet comments, its effectiveness in targeted attacks remains limited. 
Furthermore, many real-world scenarios can be viewed as gradient-free optimization problems, which have motivated attacks based on evolutionary algorithms (EA), such as STDE \cite{jiang2023efficient}. STDE utilizes spatial-temporal mutation and crossover to search for the global optimum, yet the generated patches are large, affecting the viewer's understanding of the core video content.

\subsection{Defense against Patch-based Attacks}
From the defender's perspective, several studies have proposed different defense mechanisms~\cite{naseer2019local,xiang2022patchcleanser,xu2023patchzero,lee2023defending,hwang2024temporal,song2024correction,jiang2025videopure,song2025vidtoken}. A straightforward and classical approach is adversarial training, which mixes adversarial examples with clean samples to improve model robustness. This defense remains effective when the perturbations are norm-constrained, but it performs poorly against attacks with unrestricted perturbations. Some specialized methods tailored for patch-based attacks have been proposed. Local Gradient Smoothing (LGS) \cite{naseer2019local} suppresses those high activation regions by estimating noise locations in the gradient domain and regularizing their gradients. Patch Cleanser (PC) \cite{xiang2022patchcleanser} restores benign predictions through a two-round mask approach. Additionally, due to the localized nature of patches, some object detection methods have been proposed to detect patches. For example, PatchZero \cite{xu2023patchzero} repairs adversarial examples by zeroing out detected patch regions to neutralize malicious perturbations. However, it faces two main challenges: the accuracy of patch position detection and the uncertain effectiveness of patch filling, which may even introduce new adversarial examples. Moreover, from the temporal dimension, Temporal Shuffling (TS) \cite{hwang2024temporal} mitigates the attack effectiveness by reordering the frames of an adversarial video to weaken the consistency of adversarial perturbations. 

\begin{figure*}[t]
\begin{center}
  \includegraphics[width=0.7\linewidth]{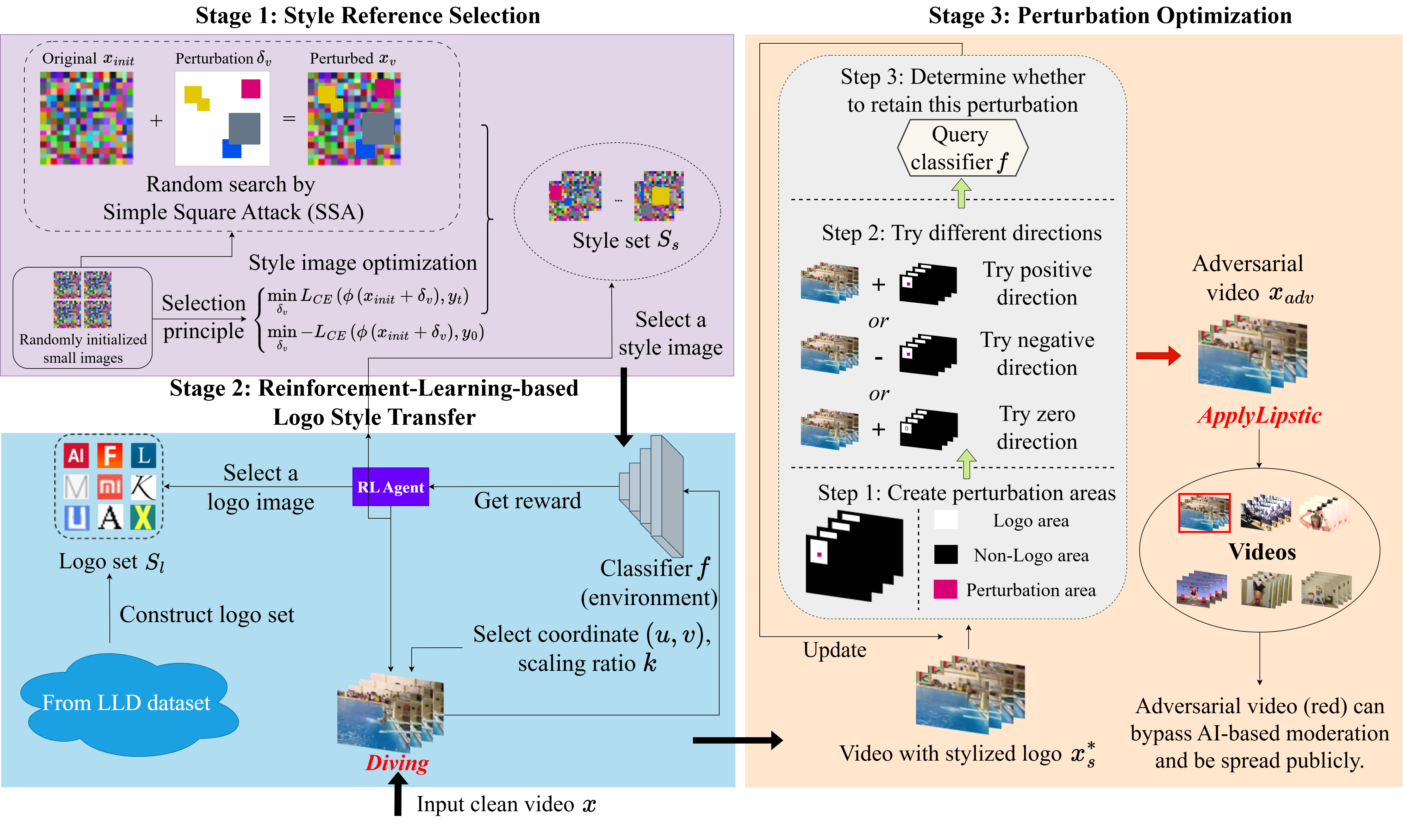}
\end{center}
\vspace{-3mm}
  \caption{Architecture of \textbf{SLA}, which includes Style Reference Selection, Reinforcement-Learning-based Logo Style Transfer and Perturbation Optimization.}
\label{fig:framework}
\end{figure*}

\section{PRELIMINARY}
\label{sec:Preliminary}
\subsection{DNN-based Video Classifier}
Many existing video classification systems in the industry predominantly use DNNs to extract features. A classifier recognizes patterns within a video containing complex semantics and provides the corresponding behavioral class labels along with confidence scores. Several representative video classifiers, such as TSN \cite{wang2018temporal}, C3D \cite{tran2015learning}, Non-local (NL) \cite{wang2018non}, TPN \cite{yang2020temporal}, and I3D \cite{carreira2017quo}, have demonstrated strong performance in video classification tasks. Specifically, TSN segments the entire video to enable the network to process longer-duration videos, thereby extracting more temporal features. C3D achieves fast inference and efficient computation through simple 3D ConvNets, making it well-suited for widespread industrial deployment. NL captures long-range dependencies among video features through non-local operations. TPN focuses on modeling visual tempos of video actions. I3D learns seamless spatio-temporal features using two-stream inflated 3D ConvNets. Among these models, C3D \cite{tran2015learning} and I3D \cite{carreira2017quo} are selected as victim models in this paper due to their representativeness and compatibility.

\subsection{Problem Definition}
Given a classifier $f$, it receives a clean video $x \in \mathbb{R}^{T \times H \times W \times C}$ as input and outputs a label $y$, where $T$, $H$, $W$, and $C$ represent the frame number, height, width, and channel number of the clean video, respectively. The attacker aims to generate an adversarial video $x_{adv}$ that causes the classifier $f$ to misclassify. 
The attacker pursues two types of objectives. As shown in Equation~\ref{eq:attack_goal}, in a targeted attack, the goal is to force the classifier to predict a specified target label $y_t$. 
In an untargeted attack, the goal is simply to change the prediction away from the original label $y_0$.
\begin{equation}\label{eq:attack_goal}
\begin{cases}f\left(x_{adv}\right)=y_t, & \text { if targeted }, \\ 
f\left(x_{adv}\right) \neq y_0, & \text { if untargeted. }\end{cases}
\end{equation}

\subsection{Threat Model}

\subsubsection{Attack Scenarios} \textbf{SLA} applies in an offline scenario, where the attacker can iteratively optimize perturbations before deployment. In addition, \textbf{SLA} focuses on one-on-one attacks, where the attacker crafts a tailored perturbation for each input video by superimposing a stylized logo onto a static clean video. The resulting adversarial video causes the victim model to produce an incorrect prediction.

\subsubsection{Attack Capabilities} 
\begin{figure}
\centering
	\captionsetup{
		font={scriptsize}, 
	}
	\begin{adjustbox}{valign=t}
		\includegraphics[width=0.9\linewidth]{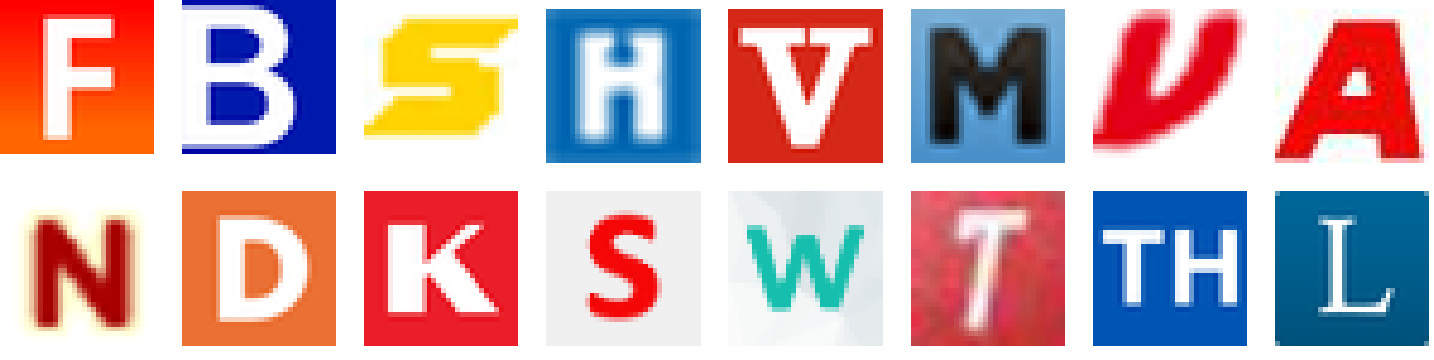}
	\end{adjustbox}
	\caption{Some examples of logo references.}
	\label{figs:Logos}
    \vspace{-3mm}
\end{figure}

Attackers can establish a suitable logo set from public datasets (\eg LLD dataset \cite{sage2018logo}) for logo reference selection. \reffig{figs:Logos} shows some examples of logo references. 
Following previous work~\cite{yang2020patchattack,guo2019simple}, we consider the practical black-box setting, where the attacker cannot access the gradients of the video classifier. 
In addition, consistent with~\cite{cao2023stylefool,cao2024logostylefool}, a maximum query limit is set to avoid meaningless queries when the algorithm struggles to converge, which is also consistent with the fact that query costs are limited in practical scenarios.

\section{METHOD}
\label{sec:METHOD}
This section introduces our proposed \textbf{SLA}, which includes three main stages: Style Reference Selection, RL-based Logo Style Transfer, and Perturbation Optimization. The attack pipeline of \textbf{SLA} is illustrated in \reffig{fig:framework}.

\subsection{Stage 1: Style Reference Selection}
The goal of this stage is to establish a style set that enables the stylized logo to carry more features about the target label in targeted attacks (other labels in untargeted attacks).
\subsubsection{Purpose of Selection}
To reduce computational complexity, we first initialize a small tensor $x_{init} \in \mathbb{R}^{\tilde{H} \times \tilde{W} \times C}$ following a uniform distribution as the candidate style reference tensor, where $\tilde{H}$ and $\tilde{W}$ represent the height and width of the initialized tensor respectively. For targeted attacks, we perform local pixel-level modifications on $x_{init}$ using a patch-based unrestricted perturbation, aiming to make $x_{init}$ carry more features of the target class. Let $x_v = x_{init}+\delta_v$, where $\delta_v$ represents the perturbation such that $f(\phi(x_v)) = y_t$, where the symbol $\phi$ denotes the resizing function, which scales up the small, perturbed initialized tensor to match the dimensions of the original input video $x$. This is because in subsequent attacks, the logo stylized by $x_v$ is attached to the clean video, which moves the new video close to the decision boundary of the classifier $f$. For untargeted attacks, a suitable $x_v$ only needs to ensure that the classification result after superimposing the perturbation $\delta_v$ is no longer the label $y_0$ of the original video, \ie $f(\phi(x_v)) \neq y_0$. Therefore, we attempt to find appropriate perturbations $\delta_v^*$ to obtain a perturbed image $x^*_v = x_{init} + \delta_v^*$ satisfying the aforementioned conditions. This can be achieved using a simple cross-entropy loss $L_{CE}$:
\begin{equation}
\left\{ \begin{array}{l}
\mathop {\min }\limits_{{\delta _v}} L_{CE}\left( {\phi \left( { x_{init}} + {\delta _v}\right) ,{y_t}} \right), ~ \rm{targeted},\\
\mathop {\min }\limits_{{\delta _v}}  - L_{CE}\left( {\phi \left( {x_{init}} + {\delta _v}\right) ,{y_0}} \right), ~ \rm{untargeted}.
\end{array} \right.
\end{equation}

\subsubsection{Simple Square Attack}
Inspired by both the Simple Black-box Attack (SimBA) \cite{guo2019simple} and Square Attack (SA) \cite{andriushchenko2020square}, we propose an improved version of SA, termed Simple Square Attack (SSA), which is also a score-based attack. Specifically, SSA aims to add multiple rounds of square-shaped perturbations to $x_{init}$. The perturbations added in the same round are identical. Each square-shaped perturbation is parameterized by its side length $Q$ and the top-left pixel coordinates $(\xi, \psi)$. The side length $Q$ is computed as
\begin{equation}
Q = round(\sqrt{\alpha\times \tilde{H} \times \tilde{W}}),
\end{equation}
where $round$ denotes rounding to the nearest integer and $\alpha$ controls the perturbation size, which follows a predefined piece-wise constant schedule that changes in stages as the number of iterations increases. The position coordinates are randomly selected in the range of $[0, \tilde{H} - Q]$ and $[0, \tilde{W} - Q]$ to ensure that the square perturbation lies entirely within the spatial boundary of $x_{init}$, \ie
\begin{equation}
(\xi, \psi) = (\texttt{rs}(0, \tilde{H}-Q) \cap \mathbb{Z}, \texttt{rs}(0, \tilde{W}-Q) \cap \mathbb{Z}),
\end{equation}
where \texttt{rs} represents randomly selecting pixel coordinates within a certain range. Then, the perturbation for each round is represented as a tensor $\delta_v \in \mathbb{R}^{Q \times Q \times C}$. All elements of $\delta_v$ are initialized to a constant value. 
The pixel values of $x_{init}$ are then updated in different directions under the influence of $\delta_v$. Specifically, in each round, every pixel of $x_{init}$ can be modified by one of $\{-\eta, 0, \eta\}$ relative to its original value, where $\eta$ denotes the step size. Perturbations that effectively mislead the classifier $f$ are retained for subsequent iterations.

To expand the search space for RL-based Logo Style Transfer in the second stage and to obtain more diverse stylized logos that can resist defenses, the aforementioned style reference selection is performed $N$ times to obtain a variety of candidates, forming the style set $S_s$. The overall process of Style Reference Selection is depicted in Algorithm \ref{alg:SA}, where $\randominit$ represents the random initialization for small images, $\alphaschedule$ is a predefined piece-wise constant schedule, $\randomselect$ represents the random selection of the coordinates, 
and $\append$ adds the suitable style image to the style set $S_s$.

\noindent
\begin{algorithm}[t]
\caption{Style Reference Selection (Targeted).}\label{alg:SA}\small
\KwIn{Classifier $f$, 
target label ${y_t}$, step size $\eta$, perturbation iterations $I$, perturbation size coefficient $\alpha$.}
\KwOut{Style set $S_s$.}
    $i \leftarrow  0$\;
    ${S_s} \leftarrow  [ ]$\;
\While {not meeting the termination condition}{
    $x_{init} \leftarrow  \randominit()$\;
    $last\_score \leftarrow  p(y_t|\phi(x_{init})) $\;
    \For {$i < I$}{
        $Q \leftarrow  \alphaschedule(\alpha,i,I,\tilde{H},\tilde{W})$\;
        $(\xi,\psi) \leftarrow  \randomselect(Q,\tilde{H},\tilde{W})$\;
        $M_s \leftarrow$ a tensor of zeros with size $(\tilde{H}, \tilde{W}, C)$\;
        $M_s[\xi:\xi+Q,\, \psi:\psi+Q,\, :] \leftarrow \eta$\;
        $x_{pos} \leftarrow x_{init} + M_s$\;
        $x_{neg} \leftarrow x_{init} - M_s$\;
        $score_{pos} \leftarrow p(y_t|\phi(x_{pos}))$\;
        $score_{neg} \leftarrow p(y_t|\phi(x_{neg}))$\;
        \If{$\max(score_{pos}, score_{neg}) > last\_score$}{
            \If{$score_{pos} > score_{neg}$}{
                $x_{init} \leftarrow x_{pos}$\;
                $last\_score \leftarrow score_{pos}$\;
            }
            \Else{
                $x_{init} \leftarrow x_{neg}$\;
                $last\_score \leftarrow score_{neg}$\;
            }
        }
        \If{$f(\phi(x_{init})) = y_t$}{
            $x_v^* \leftarrow x_{init}$\;
            $S_s.\append(x_v^*)$\;
            $break$\;
        }
    }
}
\rm{\textbf{return}} $S_s$. 
\end{algorithm}

\subsection{Stage 2: RL-based Logo Style Transfer}

In this stage, we aim to generate a stylized logo with attributes optimized by RL and impose it on the clean video.

\subsubsection{Logo Style Transfer}
We first employ style transfer to stylize a logo $l$ with a style image $x_v$, and obtain the corresponding stylized logo $l_s$ under the following constraints.

To ensure that the stylized logo $l_s$ retains the semantics of the original logo $l$, the content loss is defined as:
\begin{equation}
L_{con}\left(l, l_s\right)=\sum_i \sum_j \frac{1}{H_j W_j C_j}\left\|\vartheta_j\left(l^i\right)-\vartheta_j\left(l_s^i\right)\right\|_2^2, 
\end{equation}
where the superscript $i$ denotes the $i$-th frame of the video, 
$H_j$, $W_j$, and $C_j$ represent the height, width, and channel number of the feature maps at the $j$-th layer. $\vartheta$ signifies the feature extractor, and VGG-19~\cite{simonyan2014very} is used in this paper.

Moreover, a style loss is introduced to reduce the style difference between $l_s$ and $x_v$ :
\begin{equation}
L_{sty}\left(l_s, x_v\right)=\sum_i \sum_j \frac{1}{C_j^2}\left\|G_j(x_v)-G_j\left(l_s^i\right)\right\|_2^2,
\end{equation}
where $G$ represents the Gram matrix \cite{johnson2016perceptual} in the $j$-th layer. By introducing content loss and style loss, the video with the stylized logo is expected to approach or even cross the classifier's decision boundary. For example, the video contains more information from the target class in targeted attacks. 

Finally, a total variance loss \cite{johnson2016perceptual} is implemented to make the stylized logo appear smoother:
\begin{equation}
\begin{aligned}
& {L_{tv}}\left({l_s}\right)=\sum_i {\sum_{\hat{u}, \hat{v}}}\left(\left\|{l_s^i}(\hat{u}, \hat{v})-{l_s^i}(\hat{u}+1, \hat{v})\right\|^2\right. \\
& \left.+\left\|{l_s^i}(\hat{u}, \hat{v})-{l_s^i}(\hat{u}, \hat{v}+1)\right\|^2\right),
\end{aligned}
\end{equation}
where $(\hat{u}, \hat{v})$ represents the pixel coordinates of stylized logo. This loss serves as a regularization term in the total loss.

Thus, the objective for logo style transfer is expressed as:
\begin{equation}
{l_s^*}=\underset{l_s}{\arg \min } {\lambda_c} L_{con}\left(l, l_s\right)+{\lambda_s} {L_{sty}}\left({l_s}, {x_v}\right)+{\lambda_{tv}} {L_{tv}}\left({l_s}\right),
\end{equation}
where $\lambda_c$, $\lambda_s$, and $\lambda_{tv}$ are the weights to control each loss.

\subsubsection{Logo Attribute Selection}
Selecting an appropriate location, size, logo, and style is crucial for effective attacks and for reducing the query number required during the subsequent perturbation optimization stage. Specifically, we consider the following attributes for the logo: position within the video, size, pattern, and style reference. These attributes determine the effectiveness of the attack. The combination of these attributes constitutes the search space $\mathcal{S}$ with dimension ${D_{\mathcal{S}}}$ in RL:
\begin{equation}
 \mathcal{S} = \left( u, v, k, l_{ind}, s_{ind} \right),
\end{equation}
where $u$ and $v$ represent the pixel coordinates of the top-left corner of the logo in each frame of the video, $k$ denotes the scaling factor for logo size, $l_{ind}$ stands for the index of the selected logo from the established logo set $S_l$, and $s_{ind}$ signifies the index of the style reference selected from the style set $S_s$. Given a logo with height $h$ and width $w$, the valid ranges of pixel coordinates are defined as: $u \in \left[ 0, H - kh \right] \cap \mathbb{Z}$, $v \in \left[ 0, W - kw \right] \cap \mathbb{Z}$.

The training objective of the agent is to generate a suitable sequence of attack actions, which includes five action elements: the aforementioned $u$, $v$, $k$, $l_{ind}$, and $s_{ind}$. Once the agent is sufficiently trained, it can sample an optimal sequence of actions for superimposing the logo onto the clean video, resulting in a stylized logo that is scaled by a factor of $k$, with the logo style index being $l_{ind}$ and the style reference index being $s_{ind}$, placed at position $(u, v)$ in each frame of the video.

To minimize the impact on the core semantics of the video, the logo is encouraged to be positioned near the corners of the video and kept relatively small. More importantly, 
confidence scores are incorporated into the reward function, which helps reduce the number of queries required in the next stage. Overall, the reward function is constructed based on three factors: logo position, logo size, and confidence scores, and can be formulated as:
\begin{equation}
R = \left\{ \begin{array}{l}
\log p\left( {{y_t}|{x_s}} \right) - {\mu _a}{k^2}hw - {\mu _d}d_m, ~ \rm{targeted},\\
\log \left( {1 - p\left( {{y_0}|x_s} \right)} \right) - {\mu _a}{k^2}hw - {\mu _d}d_m, ~ \rm{untargeted},
\end{array} \right.
\end{equation}
where $x_s$ is the video with the stylized logo added, ${k^2}hw$ actually calculates the area of the scaled stylized logo, and $d_m$ denotes the minimum distance from the logo to the nearest corner of the video, \ie
\begin{equation}
d_m = \min \{d_{nw},d_{ne},d_{sw},d_{se}\},
\end{equation}
where $d_{nw}$, $d_{ne}$, $d_{sw}$, and $d_{se}$ represent the distances from the stylized logo to the northwest, northeast, southwest, and southeast corners of the video, respectively. $\mu_a$ and $\mu_d$ denote the penalty coefficients for the logo area and the distance mentioned above, respectively.

For the policy network, similar to previous work \cite{yang2020patchattack, chen2022attacking}, we choose an LSTM network with a fully connected layer. 
At each time step $t$, the agent outputs an action probability distribution 
$p\left( {{a_t}|\left( {{a_1},\; \cdots ,{a_{t - 1}}} \right)} \right)$, where the action $a_t$ is sampled from a Categorical distribution, and its corresponding sampling probability is recorded. 
After completing all time steps, the agent generates a sequence of actions, as well as their associated probabilities $\pi _{\theta _p}$, where $\theta _p$ represents the parameters of the policy network. To align the sampled action distribution with the reward, the loss function is defined as follows.
\begin{equation}
L\left( {{\theta _p}} \right) =  - {\mathbb{E}_{\tau  \sim {\pi _{{\theta _p}}}}}\left[ {R\left( \tau  \right)} \right].
\end{equation}

The parameters of the policy network are updated using the approximate gradients obtained through multiple trajectory sampling \cite{williams1992simple}, which can be expressed as:
\begin{equation}
{\nabla _{{\theta _p}}}L\left( {{\theta _p}} \right) \approx - \frac{1}{\Omega }\sum\limits_{\tau  = 1}^\Omega  {\sum\limits_{t = 1}^{D_{\mathcal{S}}} {{\nabla _{{\theta _p}}}\log {\pi _{{\theta _p}}}\left( {{a_t}|{h_t}} \right)} } R\left( \tau  \right),
\end{equation}
where $\Omega$ and $h_t$ respectively represent the number of sampling trajectories and the hidden state of the LSTM. By continuously optimizing the policy parameters $\theta _p$, the agent learns a sequence of actions that produce a logo with optimal attributes that can deceive the classifier.

\subsubsection{Benefits of Stylized Logos}
Stylized logos offer several key advantages. Firstly, stylized logos carry more information about the target class in targeted attacks or about the classes other than the original class in untargeted attacks, thereby pushing videos with stylized logos close to or even across the decision boundary of the classifier. Secondly, small and corner-located perturbations are difficult for defense mechanisms to detect, and the diversity of style images leads to a wide range of stylized logos, making it challenging for defensive methods to generalize across all variations. Finally, human observers are less likely to notice small stylized patches placed near the corners of a video, lowering visual salience and making the attack less perceptible.

\subsection{Stage 3: Perturbation Optimization}
To overcome the limitation of patch-based attacks, particularly in targeted attacks where the search space in RL is constrained, we introduce an additional stage after the previous processes. This extension, termed Logo-SSA, perturbs the region occupied by the logo. By expanding the search space, Logo-SSA can enhance the attack effectiveness for targeted attacks.

Specifically, we use random search to alter pixels within the logo region in a square-shaped manner. In each round, each pixel in the logo region can only be changed by one of \{-$\eta_p$, 0, $\eta_p$\} relative to its original value, where $\eta_p$ denotes the perturbation step size. The optimization process is then performed as follows.
\begin{equation}
\mathop {\min }\limits_{\delta_p}  L_{CE}\left( {x_s^* + M \odot \delta_p ,{y_t}} \right),\text{s.t.}{\left\| {M \odot \delta_p } \right\|_p} \le \varepsilon ,
\end{equation}
where $x_s^*$ denotes the video added with the optimal logo, $\varepsilon$ represents the perturbation threshold, $\delta_p$ denotes the perturbation, $M \in \mathbb{R}
^{T \times H \times W \times C}$ signifies the logo mask where the value in the logo region is 1 and 0 otherwise. Based on previous experimental attempts, the majority of videos cannot be turned into adversarial videos in just one round of attack; therefore, the optimization process iterates for multiple rounds until success. In this stage, the same optimization principles as in Algorithm \ref{alg:SA} are applied within the logo area.

\subsection{\textbf{SLA} Recap}
The overall process of \textbf{SLA} is shown in Algorithm \ref{alg:SLA}, where $\StyleReferenceSelection$ outputs the style set, $\logotransfer$ represents the style transfer for the logo, $\calmask$ generates the logo mask based on the position and size, $\pad$ resizes the logo and places it in the correct position in each frame of the video, with pixels of the non-logo area filled with zeros. To conclude, the main workflow of \textbf{SLA} involves three stages: constructing a suitable style set, choosing optimal logo attributes under RL guidance, and optimizing the perturbations within the logo region. For untargeted attacks, $y_t$ is replaced with $y_0$ in the algorithm.

\noindent
\begin{algorithm}[t]
\caption{Stylized Logo Attack (Targeted).}\label{alg:SLA}\small
\KwIn{Classifier $f$, clean video $x$, 
target label ${y_t}$, perturbation iterations $I$, perturbation size coefficient $\alpha$, step size $\eta$ in stage 1, logo set $S_l$, search space $\mathcal{S}$, perturbation step size $\eta_p$ in stage 3, perturbation threshold $\varepsilon$.}
\KwOut{Adversarial video $x_{adv}$.}
    $S_{s} \leftarrow \StyleReferenceSelection(f, y_t, \eta, I, \alpha)$\;
    \While {not meeting the termination condition}{
        ${a} \leftarrow$ an action sequence $\left( {u,v,k,{l_{ind}},{s_{ind}}} \right)$ sampled from $\mathcal{S}$\;
        $x_{v}^* \leftarrow  S_{s}(s_{ind})$\;
        $l_{s}^* \leftarrow  \logostyletransfer(S_{l}(l_{ind}), x_{v}^*)$\; 
        $M \leftarrow  \calmask(u,v,k,x) $\;
        $x_s \leftarrow x + M \odot \pad\left( {l_s^*} \right)$\;
        Calculate reward $R$\;
        Calculate RL loss gradient ${\nabla _{{\theta _p}}}L\left( {{\theta _p}} \right)$\;
        Update policy network and the best video $x_s^*$\;
    }
$x_{adv} \gets {\textit{Logo-SSA}} \left( f, x_s^*, y_t, u, v, \eta_p, \varepsilon \right)$\;
\rm{\textbf{return}} $x_{adv}$. 
\end{algorithm}

\section{EXPERIMENTAL EVALUATION}
\label{sec:EXPREIMENTAL EVALUATION}

\subsection{Setup}
\subsubsection{Datasets}
We select four mainstream video datasets, 
UCF-101 \cite{soomro2012ucf101}, HMDB-51 \cite{kuehne2011hmdb}, Kinetics-400 \cite{kay2017kinetics} and Kinetics-700 \cite{smaira2020short}. 
UCF-101 is an action recognition dataset featuring real-life action videos from YouTube, containing 13,320 videos in 101 action classes. HMDB-51, on the other hand, is a video database comprising 51 action categories, with approximately 7,000 manually annotated clips from various platforms, ranging from digital movies to videos on YouTube. 
Kinetics-400 is a large-scale dataset containing 400 human action classes, each with more than 400 segments extracted from real-world YouTube videos, making it two or more orders of magnitude larger than earlier benchmarks.
Kinetics-700, officially released in 2020, is an extension of Kinetics-400. It covers 700 classes and contains more than twice the data. 
In experiments, 100 videos are randomly selected from the test sets of each dataset to attack the target models. All selected videos are correctly classified by their respective classifiers. 

In addition, a logo set is constructed from the LLD dataset \cite{sage2018logo}, which includes over 600,000 logos from around the world. Certain logos featuring an extensive array of transparent pixels may exhibit undesirable effects after style transfer, such as irregular coloration against a white backdrop, thereby diminishing the stealthiness of the stylized logos. Consequently, such logos are deemed incompatible with the style transfer process. To address this, data preprocessing is conducted on LLD by automatically removing logos with transparent pixels or a large number of white pixels, resulting in a suitable logo set $S_l$. 
It is worth noting that, Kinetics-400 and Kinetics-700, with their larger number of classes and finer granularity, present greater challenges for conducting targeted attacks on models leveraging these datasets. Results of attack performance in Section \ref{sec:Attack Performance} also corroborate this observation.

\subsubsection{Target Models}
C3D \cite{tran2015learning} and I3D \cite{carreira2017quo} are selected as target models for their distinct approaches to video analysis. C3D employs 3D convolution to capture temporal features effectively, leading to robust classification accuracy. On the other hand, I3D focuses on the relationship between consecutive frames, leveraging optical flow to discern actions and achieving considerable classification performance.

For pretraining, C3D and I3D are trained on the training sets of UCF-101 and HMDB-51. For Kinetics, the weights provided by MXNet \cite{chen2015mxnet} are directly utilized. Due to differences in data processing, the input formats for C3D and I3D vary across datasets. 
Specifically, for I3D on Kinetics-400, each input consists of 32 frames of size 224$\times$224 pixels, while other inputs contain 16 frames of size 112$\times$112 pixels.
The video classification accuracies for C3D and I3D are as follows: 83.54\% and 61.70\% on UCF-101, 66.77\% and 47.92\% on HMDB-51, 59.52\% and 71.84\% on Kinetics-400, and 44.72\% and 67.65\% on Kinetics-700, respectively.

\subsubsection{Competitors}
Five patch-based attacks, Sparse-RS \cite{croce2022sparse}, Adv-watermark \cite{jia2020adv}, PatchAttack \cite{yang2020patchattack}, BSC \cite{chen2022attacking} and our conference work LogoStyleFool \cite{cao2024logostylefool}, are selected as competitors. PatchAttack is extended to videos and we consider rectangular patches with RGB perturbations to ensure fair comparison. Since BSC primarily focuses on untargeted attacks, it requires slight adjustment for targeted attacks. Both attacks employ RL iterations for patch optimization, and the iteration is stopped once the reward converges. However, success is not guaranteed, especially in targeted attacks. Due to the query constraints in our attack, the batch size and iteration steps of them are increased to achieve relative fairness. Additionally, recognizing the similarity between watermarks and logos, Adv-watermark is also applied to video attacks. The default parameter values are maintained for competitors. However, in BSC attacks, the feasibility of \textbf{SLA}'s third stage (perturbation optimization) is reduced due to varying bullet comment positions in video frames. Thus, for fair comparison, the results obtained through \textbf{SLA} without this stage are also presented. In addition, the query limit is set to $3\times10^5$ akin to existing video attacks, V-BAD \cite{jiang2019black} and StyleFool \cite{cao2023stylefool}.

\subsubsection{Metrics}
The attack performance is evaluated using the following metrics.

\noindent\textbf{Fooling Rate (FR)}: FR measures the proportion of adversarial videos that successfully mislead the classifier to either the target class (for targeted attacks) or any other class (for untargeted attacks) within the preset query limit.

\noindent\textbf{Average Query (AQ)}: AQ$_1$, AQ$_2$, and AQ$_3$ represent the average number of queries required in the first, second, and third stages, respectively. Notably, most targeted attacks require involvement of the third stage, which consumes the most queries. In contrast, untargeted attacks often succeed by the end of the second stage. Therefore, the fooling rate ($^2$FR) and the number of queries required to achieve success by the end of the second stage ($^2$AQ) are also calculated to demonstrate the excellent performance of \textbf{SLA} without the need for perturbation optimization, especially in untargeted attacks. 

\noindent\textbf{Average Occluded Area (AOA)}: AOA measures the average occluded area in the videos, which evaluates whether patches/logos significantly affect the core video semantics.

\noindent\textbf{Temporal Inconsistency (TI)}: TI is introduced to measure the temporal inconsistency of adversarial videos. Specifically, a warp function \cite{ruder2018artistic} is utilized to calculate the warping error $E_{warp}$, which is defined as follows.
\begin{equation}\label{epair}
{E_{pair}}\left( {{x_t},{x_m}} \right) = \frac{1}{{HWC}}{O_{t,m}}{\left\| {{x_t} - {\mathcal W}\left( {{x_m}} \right)} \right\|_1},
\end{equation}
\begin{equation}\label{ewarp}
{E_{warp}}\left( {{x_t}} \right) = \frac{1}{{T - 1}}\sum\nolimits_{t = 2}^T {{E_{pair}}\left( {{x_t},{x_1}} \right) + {E_{pair}}\left( {{x_t},{x_{t - 1}}} \right)},
\end{equation}
where $O_{t,m}$ denotes the occlusion mask matrix corresponding to the $t$-th and $m$-th frame $x_t$ and $x_m$, and $\mathcal W$ signifies the operation of backward warping, which utilizes optical flow to map pixels from one frame to another.

\subsubsection{Hyperparameter Search}
We set the three hyperparameters, style image number $N_{s}$, logo number $N_{l}$, and step size $\eta_p$ to 5, 80 and 0.2, respectively, after conducting a comprehensive grid search. More details are provided in the supplement.

\begin{figure}[t]
\centering
	\captionsetup{
		font={scriptsize}, 
	}
	\begin{adjustbox}{valign=t}
		\includegraphics[width=0.8\linewidth]{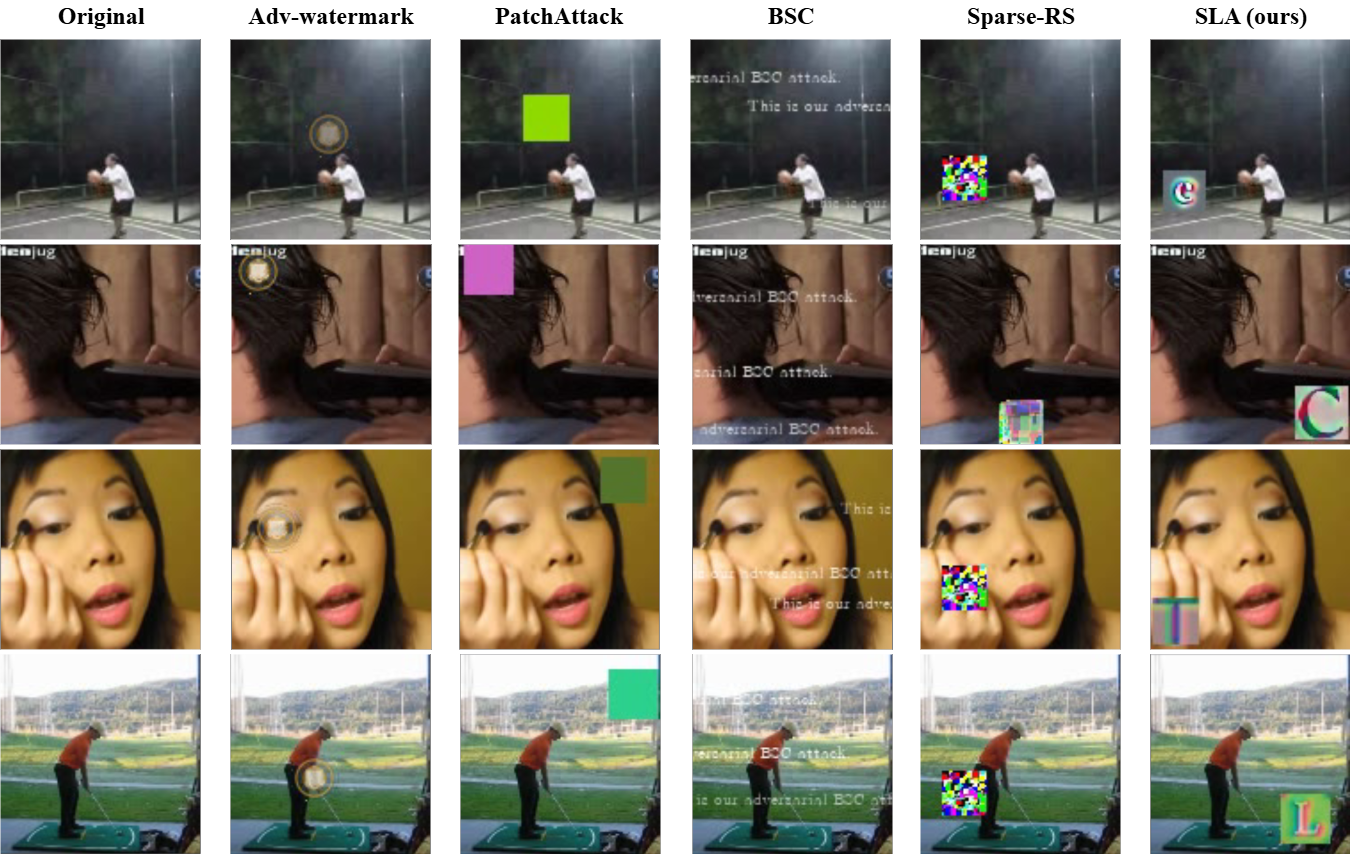}
	\end{adjustbox}
	\caption{Examples of different attacks.}
	\label{figs:different_attacks}
    \vspace{-4mm}
\end{figure}

\subsection{Attack Performance}
\label{sec:Attack Performance}
\subsubsection{Qualitative Analysis}

\reffig{figs:different_attacks} displays the adversarial samples generated by different methods. PatchAttack \cite{yang2020patchattack} and Sparse-RS \cite{croce2022sparse} generate easily detectable patches due to their lack of semantic nature. A significant drawback of Adv-watermark \cite{jia2020adv} is that it does not utilize RL to build the connection between the patch and its positions. Therefore, the patch can affect the core semantics of the video, which is markedly different from benign watermarks commonly added in daily scenarios.
As will be discussed later, 
BSC \cite{chen2022attacking}
faces the challenge of limited search space, which seriously affects the performance of targeted attacks. 
In addition, many video websites, such as YouTube and Bilibili, have bullet-screen shutdown functions, in which case the anomalies of BSC's adversarial samples will be highlighted. In contrast, in \textbf{SLA}, the stylized logos are located in the corners of the video, which means that the main semantics of the original video are not affected. Moreover, the natural semantics of the logo make it difficult for the patch to be detected when the current video is under attack. 
Figures S.1 and S.2 in the supplement show more adversarial examples of \textbf{SLA} in targeted and untargeted attacks, respectively. 
In particular, we show that different stylized logos can mislead target models into producing various predictions. Furthermore, the comparison of clean and adversarial video frames in targeted and untargeted scenarios is presented in Figures S.3, S.4 and S.5 in the supplement.

\begin{table*}[ht]
\centering
\caption{Attack performance comparison on UCF-101. 
Bold denotes the best results when FR exceeds 50\% (targeted) or 80\% (untargeted).
}
\label{tab:attack_performance_ucf101}
\footnotesize
\resizebox{0.7\linewidth}{!}{
\begin{tabular}{ccrrrrrrrrrrrr}
\toprule
\multirow{2}{*}[-0.5ex]{Model} & 
\multirow{2}{*}[-0.5ex]{Attack} & \multicolumn{4}{c}{UCF-101-Targeted} &  \multicolumn{4}{c}{UCF-101-Untargeted} \\
\cmidrule(r){3-6}\cmidrule(r){7-10}
& & FR($^2$FR)$\uparrow$ & AQ($^2$AQ)$\downarrow$ & AOA$\downarrow$ & TI$\downarrow$ & FR($^2$FR)$\uparrow$ & AQ($^2$AQ)$\downarrow$ & AOA$\downarrow$ & TI$\downarrow$ \\
\midrule
\multirow{6}{*}{C3D} 
& Sparse-RS \cite{croce2022sparse} & 57\% & 14,013.2 & 4.98\% & 4.98 &   \textbf{98\%} & \textbf{213.8} & \textbf{4.98\%} & 6.11  \\
& Adv-watermark \cite{jia2020adv} & 2\% & 1,259.2 & 6.06\% & 4.66  & 69\% &  670.6 & 5.37\% & 3.73  \\
& PatchAttack \cite{yang2020patchattack} & 5\% & 33,020.7& 5.11\% & 75.62 & 80\% & 7,844.8 & 6.94\% & 102.06  \\
& BSC \cite{chen2022attacking} & 8\% & 27,608.4 & 7.15\% & 3.80 & 94\% & 2,228.1 & 7.26\% & 3.89 \\
& LogoStyleFool \cite{cao2024logostylefool} & 44\%(8\%) & 22,993.9(4,124.4) & 5.51\% & 3.76  & 94\%(79\%) & 4,661.5(978.4) & 6.52\% & 3.67  \\
& \textbf{SLA} & \textbf{65\%}(11\%) & \textbf{9,288.2}(3,446.7)& \textbf{4.88\%} & \textbf{4.38} & 96\%(80\%)& 3,313.6(782.7) & 5.24\% & \textbf{3.66} \\
\midrule
\multirow{6}{*}{I3D} 
& Sparse-RS \cite{croce2022sparse} & 53\% & 13,961.3 & \textbf{4.98\%} & 5.10  & 88\% &  \textbf{933.8} & \textbf{4.98\%} & 6.88 \\
& Adv-watermark \cite{jia2020adv} & 4\% & 1,417.5 & 4.56\% & 4.59 &  56\% & 932.7 & 4.67\% & 3.46  \\
& PatchAttack \cite{yang2020patchattack} & 7\% & 32,604.8 & 4.67\% & 51.49 & 89\% & 4,082.1 & 5.97\% & 11.93 \\
& BSC  \cite{chen2022attacking} & 9\% & 27,672.2 & 7.02\% & 3.68  & 81\% & 2,087.5 & 6.18\% & \textbf{3.87} \\
& LogoStyleFool \cite{cao2024logostylefool} & 37\%(6\%) & 20,315.5(2,999.2) & 5.47\% & 3.92 & 95\%(83\%) & 3,955.3(889.7) & 5.45\% & 4.01  \\
& \textbf{SLA} & \textbf{55\%}(8\%) & \textbf{10,443.7}(2,732.1)& 5.34\% & \textbf{4.28} & \textbf{97\%}(88\%)& 3,089.6(794.1) & 5.39\% & 3.97  \\
\bottomrule
\end{tabular}
}
\end{table*}

\begin{table*}[ht]
\centering
\caption{Attack performance comparison on HMDB-51. 
Bold denotes the best results when FR exceeds 50\% (targeted) or 80\% (untargeted).}
\label{tab:attack_performance_hmdb51}
\footnotesize
\resizebox{0.7\linewidth}{!}{
\begin{tabular}{ccrrrrrrrrrrrr}
\toprule
\multirow{2}{*}[-0.5ex]{Model} & 
\multirow{2}{*}[-0.5ex]{Attack} & \multicolumn{4}{c}{HMDB-51-Targeted} &  \multicolumn{4}{c}{HMDB-51-Untargeted} \\
\cmidrule(r){3-6}\cmidrule(r){7-10}
& & FR($^2$FR)$\uparrow$ & AQ($^2$AQ)$\downarrow$ & AOA$\downarrow$ & TI$\downarrow$ & FR($^2$FR)$\uparrow$ & AQ($^2$AQ)$\downarrow$ & AOA$\downarrow$ & TI$\downarrow$ \\
\midrule
\multirow{6}{*}{C3D} 
& Sparse-RS \cite{croce2022sparse} & 71\% & 11,044.9 & \textbf{4.98\%} & 4.78  & 97\% &  \textbf{764.1} & \textbf{4.98\%} & 5.31\\
& Adv-watermark \cite{jia2020adv} & 3\% & 1,038.1 & 4.24\% & 5.14 &  73\% &  610.2 & 4.35\% & 2.69\\
& PatchAttack \cite{yang2020patchattack} &  4\% & 28,829.1& 5.17\% & 42.03 & 88\% & 1,934.3 & 6.01\% & 66.33 \\
& BSC \cite{chen2022attacking} & 25\% & 22,684.7 & 6.12\% & 4.95 & 97\% & 1,739.4 & 7.85\% & 3.50 \\
& LogoStyleFool \cite{cao2024logostylefool} & 46\%(12\%) & 18,200.9(3,161.9) & 5.14\% & 3.15 & 96\%(86\%) & 3,396.6(901.6) & 5.80\% & \textbf{3.03} \\
& \textbf{SLA} & \textbf{76\%}(14\%) & \textbf{7,124.1}(2,854.6)& 5.01\% & \textbf{3.67}  & \textbf{98\%}(88\%)& 1,624.3(693.7) & 5.59\% & 3.14 \\
\midrule
\multirow{6}{*}{I3D} 
& Sparse-RS \cite{croce2022sparse} & \textbf{74\%} & \textbf{8,217.0} & \textbf{4.98\%} & 5.44 & 92\% &  2,010.5 & \textbf{4.98\%} & 5.56  \\
& Adv-watermark \cite{jia2020adv} & 4\% & 1,539.3 & 5.44\% & 4.01 & 78\% & 515.2 & 5.16\% & 3.65\\
& PatchAttack \cite{yang2020patchattack} & 12\% & 26,431.6 & 4.72\% & 303.34 & 89\% & 3,436.4 & 5.42\% & 15.79\\
& BSC \cite{chen2022attacking} & 13\% & 26,311.3 & 6.19\% & 4.85 & 90\% & 3,137.5 & 6.89\% & 4.57 \\
& LogoStyleFool \cite{cao2024logostylefool} & 43\%(11\%) & 20,612.8(4,723.9) & 5.85\% & 3.52 & 94\%(88\%) & 849.5(774.6) & 5.48\% & 3.43 \\
& \textbf{SLA} & 66\%(13\%) & 11,509.6(3,368.1)& 6.08\% & \textbf{3.46}  & \textbf{100\%}(96\%) & \textbf{624.9}(614.9)& 5.17\% & \textbf{3.40} \\
\bottomrule
\end{tabular}
}
\end{table*}

\begin{table*}[ht]
\centering
\caption{Attack performance comparison on Kinetics-400. Bold denotes the best results when FR exceeds 30\% (targeted) or 80\% (untargeted).}
\footnotesize
\resizebox{0.7\linewidth}{!}{
\begin{tabular}{ccrrrrrrrrrrrr}
\toprule
\multirow{2}{*}[-0.5ex]{Model} & 
\multirow{2}{*}[-0.5ex]{Attack} & \multicolumn{4}{c}{Kinetics-400-Targeted} &  \multicolumn{4}{c}{Kinetics-400-Untargeted} \\
\cmidrule(r){3-6}\cmidrule(r){7-10}
& & FR($^2$FR)$\uparrow$ & AQ($^2$AQ)$\downarrow$ & AOA$\downarrow$ & TI$\downarrow$ & FR($^2$FR)$\uparrow$ & AQ($^2$AQ)$\downarrow$ & AOA$\downarrow$ & TI$\downarrow$ \\
\midrule
\multirow{6}{*}{C3D} 
& Sparse-RS \cite{croce2022sparse} & \textbf{59\%} & \textbf{14,971.2} & 4.98\% & 7.75 & 64\% &  197.5 & 4.98\% & 11.56 \\
& Adv-watermark \cite{jia2020adv} & 2\% & 2,062.1 & 6.41\% & 5.94 & 75\% &  604.4 & 4.90\% & 5.02 \\
& PatchAttack \cite{yang2020patchattack} & 3\% & 25,121.1& 4.64\% & 25.57  & 86\% & 4,334.4 & 6.78\% & 19.71 \\
& BSC \cite{chen2022attacking} & 3\% & 29,124.8 & 6.58\% & 5.08 & 84\% & 2,217.6 & 6.70\% & 5.61 \\
& LogoStyleFool \cite{cao2024logostylefool} & 22\%(4\%) & 21,093.3(7,559.1) & 4.91\% & 5.39 & 91\%(82\%) & 1,824.9(930.1) & \textbf{5.91\%} & 5.44 \\
& \textbf{SLA} & 39\%(5\%) & 17,641.3(5,601.4)& \textbf{4.42\%} & \textbf{5.01} & \textbf{97\%}(87\%)& \textbf{1,552.4}(900.8) & 6.18\% & \textbf{4.33} \\
\midrule
\multirow{6}{*}{I3D} 
& Sparse-RS \cite{croce2022sparse} & 31\% & 22,723.9 & 4.98\% & 6.93 & 79\% &  4,531.0 & 4.98\% & 9.63\\
& Adv-watermark \cite{jia2020adv} & 3\% & 1,866.9 & 4.79\% & 5.29 & 66\% & 780.5 & 4.45\% & 5.54 \\
& PatchAttack \cite{yang2020patchattack} & 5\% & 21,585.4 & 4.77\% & 12.47 & 81\% & 5,839.9 & 7.31\% & 141.81 \\
& BSC \cite{chen2022attacking} & 6\% & 29,409.4 & 7.12\% & 5.58 & 76\% & 2,874.0 & 7.70\% & 5.65 \\
& LogoStyleFool \cite{cao2024logostylefool} & 20\%(6\%) & 23,602.7(6,221.6) & 4.62\% & 5.44 & 83\%(82\%) & 4,422.4(1,471.4) & \textbf{5.76\%} & 5.69 \\
& \textbf{SLA} & \textbf{33\%}(7\%) & \textbf{10,921.8}(5,283.5)& \textbf{3.77\%} & \textbf{5.21} & \textbf{95\%}(83\%) & \textbf{3,679.9}(1,224.3)& 5.82\% & \textbf{5.49} \\
\bottomrule
\end{tabular}
}
\label{tab:attack_performance_Kinetics400}
\end{table*}

\begin{table*}[ht]
\centering
\caption{Attack performance comparison on Kinetics-700. Bold denotes the best results when FR exceeds 30\% (targeted) or 80\% (untargeted).}
\footnotesize
\resizebox{0.7\linewidth}{!}{
\begin{tabular}{ccrrrrrrrrrrrr}
\toprule
\multirow{2}{*}[-0.5ex]{Model} & 
\multirow{2}{*}[-0.5ex]{Attack} & \multicolumn{4}{c}{Kinetics-700-Targeted} &  \multicolumn{4}{c}{Kinetics-700-Untargeted} \\
\cmidrule(r){3-6}\cmidrule(r){7-10}
& & FR($^2$FR)$\uparrow$ & AQ($^2$AQ)$\downarrow$ & AOA$\downarrow$ & TI$\downarrow$ & FR($^2$FR)$\uparrow$ & AQ($^2$AQ)$\downarrow$ & AOA$\downarrow$ & TI$\downarrow$ \\
\midrule
\multirow{6}{*}{C3D} 
& Sparse-RS \cite{croce2022sparse} & 54\% & \textbf{13,421.7} & 4.98\% & 6.11 & 69\% & 192.3 & 4.98\% & 9.87 \\
& Adv-watermark \cite{jia2020adv} & 3\% & 1,563.9 & 5.78\% & 4.89  & 79\% & 548.5 & 4.12\% & 4.33\\
& PatchAttack \cite{yang2020patchattack} & 5\% & 23,469.8 & 4.08\% & 21.05 & 89\% & 4,131.4 & 5.99\% & 16.92\\
& BSC \cite{chen2022attacking} & 2\% & 28,303.9 & 5.87\% & 4.52 & 87\% & 2,082.1 & 6.18\% & 4.97 \\
& LogoStyleFool \cite{cao2024logostylefool} & 24\%(5\%) & 21,733.1(7,605.9) & 4.55\% & 4.91 & 92\%(84\%) & 1,887.2(943.7) & \textbf{5.59\%} & 5.08 \\
& \textbf{SLA} & \textbf{54\%}(6\%) & 18,053.8(5,735.5)& \textbf{3.89\%} & \textbf{4.63}  & \textbf{98\%}(89\%)& \textbf{1,615.9}(931.8) & 5.87\% & \textbf{4.05} \\
\midrule
\multirow{6}{*}{I3D} 
& Sparse-RS \cite{croce2022sparse} & 34\% & 23,556.7 & 4.98\% & 6.55 &  81\% & 5,117.1 & \textbf{4.98\%} & 9.11 \\
& Adv-watermark \cite{jia2020adv} & 4\% & 1,930.8 & 4.55\% & 5.01 & 68\% & 804.3 & 4.21\% & 5.28 \\
& PatchAttack \cite{yang2020patchattack} & 6\% & 22,491.8 & 4.53\% & 11.73 & 83\% & 6,147.6 & 7.05\% & 134.55 \\
& BSC \cite{chen2022attacking} & 5\% & 31,927.3 & 6.88\% & 5.30 & 78\% & 3,031.5 & 7.44\% & 5.41 \\
& LogoStyleFool \cite{cao2024logostylefool} & 22\%(7\%) & 24,652.4(6,488.2) & 4.38\% & 5.18 & 85\%(83\%) & 4,618.5(1,527.0) & 5.61\% & 5.52 \\
& \textbf{SLA} & \textbf{35\%}(10\%) & \textbf{11,390.7}(5,499.0)& \textbf{3.61\%} & \textbf{4.95} & \textbf{96\%}(85\%)& \textbf{3,862.5}(1,271.3)& 5.67\% & \textbf{5.31}\\
\bottomrule
\end{tabular}
}
\vspace{-3mm}
\label{tab:attack_performance_Kinetics700}
\end{table*}

\subsubsection{Quantitative Analysis}
\label{sec:Quantitative}

Tables~\ref{tab:attack_performance_ucf101},~\ref{tab:attack_performance_hmdb51},~\ref{tab:attack_performance_Kinetics400} and~\ref{tab:attack_performance_Kinetics700}
report the numerical results of the attack performance of the different methods on the four datasets.

\noindent\textbf{Comparison of FR and Queries.}
In targeted attacks, in terms of FR, Adv-watermark exhibits the worst results (always below 5\%), followed closely by PatchAttack. One possible reason for Adv-watermark's poor performance is its excessive focus on stealthiness, such as the transparency of watermarks. BSC performs relatively better, especially on HMDB-51 (25\%), which contains the fewest classes.
In contrast, the larger label spaces of Kinetics-400/700 make targeted attacks more challenging, which is also observed in Sparse-RS, LogoStyleFool, and \textbf{SLA}. Among all competitors, Sparse-RS shows the closest competitiveness to \textbf{SLA} in FR, as both adopt derivative-free random search optimization.
However, its patches lack semantic meaning and are easily detectable by humans. 
Moreover, without covert mechanisms such as the stylization in \textbf{SLA}, Sparse-RS produces higher-frequency perturbations that are more vulnerable to defenses like LGS (see Section~\ref{sec:Countermeasure}).

As for query consumption, PatchAttack and BSC require the most queries. Both methods require more than 20,000 queries in all datasets.
Adv-watermark utilizes the fewest number of queries but suffers from a low FR. In most cases, Sparse-RS does not require fewer queries compared to our method. When compared to LogoStyleFool, \textbf{SLA} still manages to achieve significant improvements in both query and FR across all scenarios. 
The possible reasons are as follows. 
First, the pixel region modified per round in SSA, which is used in \textbf{SLA}, is larger than that in SimBA, which is adopted by LogoStyleFool.
Furthermore, SSA efficiently identifies effective candidate perturbations through random search. Most importantly, \textbf{SLA} benefits from better initialization of style images and stylized logos in its first two stages, which can reduce AQ
and enhance attack effectiveness through post-RL perturbation optimization.

In untargeted attacks, PatchAttack is the most query-consuming method.
Although Adv-watermark requires fewer queries, 
its FR is at least 10\% lower than that of \textbf{SLA} in all scenarios. Although Sparse-RS demonstrates only a slight advantage over \textbf{SLA} in terms of attack performance when attacking C3D on UCF-101, it shows instability on more challenging datasets, Kinetics-400 and Kinetics-700, where its FR is about 20\% lower compared to \textbf{SLA}. BSC shows moderate performance in both FR and AQ. \textbf{SLA} demonstrates its superiority over LogoStyleFool in FR in all scenarios, owing to the higher efficiency of SSA's random search strategy. 
In addition, \textbf{SLA} can achieve an FR exceeding 80\% within just the first two stages, making it a viable variant for scenarios with more limited query budgets. 

\noindent\textbf{Comparison of AOA and TI.} 
The $AOA$ of Sparse-RS is fixed since the patch size is pre-specified. 
Adv-watermark exhibits a low $AOA$ in most cases, but its low FR limits this benefit. 
Attributed to the properties of bullet-screen comments, BSC attains the highest $AOA$ in all scenarios, typically exceeding 6\%. 
Compared with PatchAttack and BSC, \textbf{SLA} achieves a smaller $AOA$ while maintaining a high FR.

It is clear that PatchAttack lacks temporal consistency, likely due to the use of solid-color patches. Sparse-RS exhibits the same deficiency, especially on high-resolution datasets such as Kinetics-400, where high-frequency perturbations disrupt video continuity. In contrast, Adv-watermark and BSC maintain better temporal consistency by incorporating transparency and semantic design into their patches. 
LogoStyleFool and \textbf{SLA}, which superimpose stylized patches onto clean videos, exhibit strong temporal stability.
In most cases, these methods achieve the lowest $TI$ values, indicating that the generated adversarial videos retain the highest fluency.

\subsection{Attack Visualization}
\begin{figure*}[t]
\begin{center}
  \includegraphics[width=0.82\linewidth]{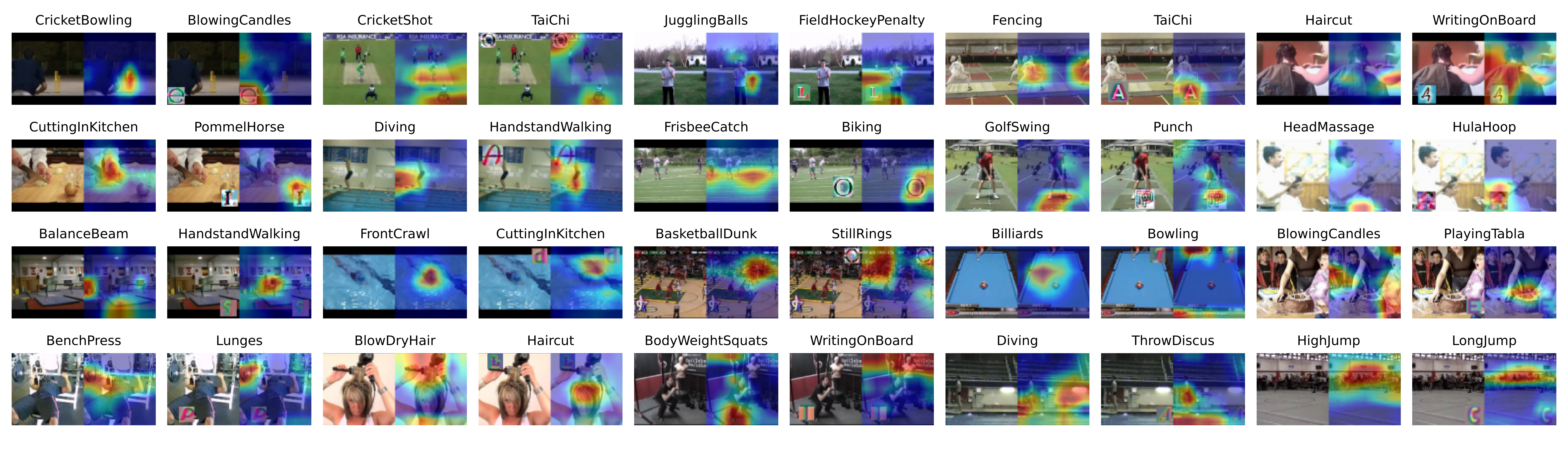}
\end{center}
\vspace{-4mm}
  \caption{Grad-CAM visualizations of \textbf{SLA}. Top two rows: targeted attacks, bottom two rows: untargeted attacks. 
  Each small figure contains a clean frame, an adversarial frame and their corresponding regions of interest. 
  The model prediction is labeled above each small figure. 
}
\label{figs:attack_visual}
\end{figure*}

To display the impact of \textbf{SLA} on the classifier, Grad-CAM \cite{selvaraju2017grad} is used to visualize the regions of interest in the target model.
\reffig{figs:attack_visual} shows the visualization results. 
Visually, the adversarial examples generated by \textbf{SLA} exhibit significant power to distort model judgments. 
Furthermore, the model's attention may not be confined to the stylized logo region and can be drawn to other areas of the video, especially on untargeted attacks. One possible reason is that perturbations will not be updated once the video becomes adversarial.

\subsection{Countermeasure}
\label{sec:Countermeasure}

In this subsection, two SOTA patch-based defenses (LGS~\cite{naseer2019local} and PC~\cite{xiang2022patchcleanser}) and one temporal defense (TS~\cite{hwang2024temporal}) are selected to evaluate \textbf{SLA}'s ability to counter these defenses. LGS regularizes gradients in the estimated perturbation region before feeding the video to the model for inference. PC aims to identify a pair of benign patches that conceal adversarial videos in a two-round masking manner. TS restores correct classification of adversarially perturbed videos by swapping frames along the temporal dimension. LGS and PC are extended to videos. FR is employed to assess the robustness of different attack methods against adversarial defenses.

Results in \reftable{tab:defense_performance} indicate that \textbf{SLA} exhibits the strongest ability to resist defenses. 
Other attacks, such as Adv-watermark, BSC, and PatchAttack, also demonstrate a certain degree of robustness, mainly because they do not include perturbation optimization, which results in smoother perturbations but poor performance in targeted attacks. 
In contrast, \textbf{SLA} introduces square-shaped perturbations to the logo area in the perturbation optimization stage, which slightly alters the logo content, as shown in Figure S.6 in the supplement. 
Even so, \textbf{SLA} still maintains strong defense resistance in most cases.
Additionally, introducing perturbation optimization into BSC or PatchAttack could trigger human detection alarms, since the bullet-screen comments or patch areas would become more conspicuous. Moreover, since the same perturbations are applied across frames in \textbf{SLA}, simple frame-shuffling defenses such as TS are less effective. Overall, current defense methods still remain far from being effective and practically deployable.

\begin{table}[t]
\centering
\caption{Robustness of attacks against defenses (metric: Fooling Rate ($\uparrow$)). The fooling rate is averaged on both targeted and untargeted attacks.}
\footnotesize
\resizebox{0.9\linewidth}{!}{
\begin{tabular}{ccccccccccc}
\toprule
\multirow{2}{*}[-0.5ex]{Model} & 
\multirow{2}{*}[-0.5ex]{Attack} & \multicolumn{3}{c}{UCF-101} & \multicolumn{3}{c}{HMDB-51} & \multicolumn{3}{c}{Kinetics-400} \\
\cmidrule(r){3-5}\cmidrule(r){6-8}\cmidrule(r){9-11}
& & LGS & PC & TS & LGS & PC & TS & LGS & PC & TS\\
\midrule
\multirow{5}{*}{C3D} 
& Sparse-RS \cite{croce2022sparse} & 33.0\% & 48.5\%& 37.5\% & 28.0\% & 34.0\% & 40.5\%& 31.0\% & 39.5\% & 31.5\%\\
& Adv-watermark \cite{jia2020adv} & 34.0\% & 41.0\% & \textbf{58.5\%} & 32.0\% & 37.0\% & \textbf{60.0\%} & 41.0\% & 36.5\% & 60.0\%\\
& PatchAttack \cite{yang2020patchattack} & 27.0\% & 39.0\% & 45.0\%& 49.5\% & 41.5\% & 51.5\% &\textbf{56.0\%} & 44.0\% & 49.0\%\\
& BSC \cite{chen2022attacking} & 30.0\% & 42.5\% & 36.5\% & 52.0\% & 41.5\% & 46.0\% & 44.5\% & 42.0\% & 37.5\% \\
& \textbf{SLA}  & \textbf{52.5\%} & \textbf{55.5\%} & 57.0\% & \textbf{56.0\%} & \textbf{45.0\%} & 54.5\% & 52.5\% & \textbf{46.5\%} & \textbf{62.5\%}\\
\midrule
\multirow{5}{*}{I3D} 
& Sparse-RS \cite{croce2022sparse} & 32.0\% & 36.5\% & 41.0\% & 26.0\% & 39.5\% & 43.0\% & 22.0\% & 50.5\% & 33.0\%\\
& Adv-watermark \cite{jia2020adv} & 36.0\% & 34.0\% & 52.0\% & 37.0\% & 58.0\% & 59.5\% & 32.0\% & 39.0\% & 58.0\%\\
& PatchAttack \cite{yang2020patchattack} & 36.5\% & 35.0\% & 33.5\% & 47.5\% & 48.5\% & 38.0\%& 43.0\% & 40.0\% & 32.0\% \\
& BSC \cite{chen2022attacking} & 40.5\% & \textbf{47.5\%} & 44.5\% & 49.0\% & 55.5\% & 52.5\% & 44.0\% & 42.5\% & 41.0\% \\
& \textbf{SLA} & \textbf{42.5\%} & 42.0\% & \textbf{59.0\%} & \textbf{52.5\%} & \textbf{59.5\%} & \textbf{61.0\%} & \textbf{45.0\%} & \textbf{61.0\%} & \textbf{59.0\%}\\
\bottomrule
\end{tabular}
}
\label{tab:defense_performance}
\vspace{-3mm}
\end{table}

\begin{table*}[tbp]  
\centering
\caption{Ablation results of \textbf{SLA}.}
\resizebox{0.8\linewidth}{!}{
\begin{tabular}{ccrrrrrrrrrrrrrrrr}
\toprule
\multirow{2}{*}[-0.5ex]{Attack scenario} & \multicolumn{5}{c}{UCF-101-Targeted} & \multicolumn{5}{c}{UCF-101-Untargeted} \\
\cmidrule(r){2-6}\cmidrule(r){7-11}
& FR($^2$FR)$\uparrow$ & AQ$_1$ & AQ$_2$ & AQ$_3$ & AQ($^2$AQ)$\downarrow$ & FR($^2$FR)$\uparrow$ & AQ$_1$ & AQ$_2$ & AQ$_3$ & AQ($^2$AQ)$\downarrow$ \\
\midrule
Random style image & 31\%(6\%) & 0 & 2,185.4 & 13,786.8 & 15,972.2(2,185.4) & 92\%(74\%) & 0 & 682.3 & 2,055.7 & 2,738.0(682.3) \\
Solid color initialization & 36\%(6\%) & 2,109.9 & 2,056.6 & 9,802.1 & 13,968.6(4,166.5) & 87\%(72\%) & 1.1 & 1,062.6 & 4,587.9 & 5,651.6(1,063.7) \\
Vertical strips initialization & 49\%(9\%) & 2,317.0 & 2,074.6 & 9,019.2 & 13,410.8(4,391.6) & 94\%(86\%) & 1.1 & 784.1 & 2,218.2 & 3,003.4(785.2) \\
\midrule
$\mu _a = 0$ & 58\%(9\%) & 2,079.4 & 1,839.0 & 8,017.5 & 11,935.9(3,918.4) & 90\%(82\%) & 1.0 & 916.3 & 3,305.4 & 4,222.7(917.3) \\
$\mu _d = 0$ & 62\%(11\%) & 2,061.6 & 1,715.7 & 7,030.7 & 10,808.0(3,777.3) & 90\%(80\%) & 1.0 & 800.1 & 3,315.1 & 4,116.2(801.1) \\
\midrule
Random step direction & 57\%(7\%) & 1,821.7 & 2,022.3 & 9,025.9 & 12,869.9(3,844.0) & 91\%(82\%) & 1.2 & 1,020.9 & 3,299.8 & 4,321.9(1,022.1) \\
\bottomrule
\end{tabular}
}
\vspace{-3mm}
\label{tab:ablation}
\end{table*}

\subsection{Ablation Study}
To validate the effectiveness of different components in \textbf{SLA}, an ablation study is conducted on the UCF-101 \cite{soomro2012ucf101} dataset, considering six variations (abbreviated as $A$ to $F$) in three stages. These variations are described as follows. 
\emph{A: Random style image (Stage 1):} Random style images are used, thus skipping Stage 1. \emph{B: Solid color initialization (Stage 1):} Solid colors are adopted instead of random initialization. \emph{C: Vertical strip initialization (Stage 1):} Vertical strips of width 1 are utilized, and the color of each strip is uniformly sampled from $\{-\eta, +\eta \}$. 
\emph{D: $\mu _a = 0$ (Stage 2):} The constraint on the size of the logo is removed. \emph{E: $\mu _d = 0$ (Stage 2):} The constraint on the distance of the logo from the corners is removed. \emph{F: Random step direction (Stage 3):} In each round, each pixel is modified by one of \{-$\eta_p$, 0, $\eta_p$\} rather than along a score-guided direction. \reftable{tab:ablation} reports the results.

\noindent\textbf{Impact of Style Reference Selection (Stage 1)}. 
Although variant A saves the required queries for selecting style images, its inability to convey target-class information causes a sharp decrease in targeted attack efficiency (\eg AQ$_3$ exceeds 13,000). On the other hand, thanks to the ability of the roughly randomized style images to mislead the classifier, the efficiency of untargeted attacks is improved. Therefore, variant A can serve as an improved version of \textbf{SLA} for untargeted attacks. Variants B and C mainly focus on exploring different initialization strategies for style images. Similar to variant A, they both perform poorly in targeted attacks. Specifically, the monotonous color from variant B makes perturbation optimization more difficult, resulting in more than 9,800 (4,500) queries in Stage 3 for targeted (untargeted) attacks. For variant C, although some studies \cite{yin2019fourier,andriushchenko2020square} in the field of image processing indicate that CNNs have shown particular sensitivity to strip-based perturbations, vertical strips are ineffective in attacks. One possible reason is the increased dimensionality from images to videos. 

\noindent\textbf{Impact of RL-based Logo Style Transfer (Stage 2)}. In theory, variants D and E directly affect the core semantics of the video, reducing its naturalness and potentially drawing human attention, as the size and position of the superimposed logo are not constrained. We define the average logo area as $\bar a = \mathbb{E}_{i}[k_i^2hw]$ and the average minimum distance to the corners as $\bar d_m = \mathbb{E}_{i}[d_{m,i}] $, where $i$ denotes the index of each adversarial video, and the averages are taken over all adversarial videos. We find that these averages increase significantly in variants D and E. Specifically, compared with $\bar a$ of 853.0 (657.3) and $\bar d_m$ of 8.8 (14.8) for targeted (untargeted) attacks in \textbf{SLA}, $\bar a$ becomes 917.1 (731.3) for targeted (untargeted) attacks in variant D, and $\bar d_m$ becomes 35.4 (23.2) for targeted (untargeted) attacks in variant E. Since both variants can make the logo conspicuous and affect the video semantics, they are not adopted, despite achieving relatively good attack performance.

\noindent\textbf{Impact of Perturbation Optimization (Stage 3)}. Variant F results in disordered optimization, causing a sharp increase in the required queries during Stage 3. 
Specifically, 
in targeted attacks, the AQ$_3$ (9,025.9) in variant F is 54.5\% higher than that in \textbf{SLA} (5,841.5). In untargeted attacks, the AQ$_3$ (3,299.8) in variant F increases by more than 30.4\% compared to that in \textbf{SLA} (2,530.9).

\subsection{User Study}
\label{sec:User Study}
To verify the naturalness of the adversarial videos generated by \textbf{SLA} and to assess whether they affect human understanding of video content, two sets of user studies are conducted to evaluate the naturalness and semanticity.  

\subsubsection{Preparation}
30 videos (10 each from UCF-101, HMDB-51, and Kinetics-400) are randomly selected for the naturalness and semanticity tests, respectively. Both sets consist of 15 clean videos and 15 adversarial videos. 
Amazon Mechanical Turk (AMT), a crowdsourcing platform operated by Amazon, was used to conduct the online survey.
We recruited 100 anonymous participants who were all over 18 years old, with English as their native language and with an approval rate of at least 95\% for their previous responses. They were evenly divided into two groups for the two tests to avoid bias.

\subsubsection{Survey Design}
All survey questions were multiple-choice with five options, and participants could select only one answer per question. 
In the naturalness test, after watching each video, participants were required to evaluate the video naturalness based on a Likert scale~\cite{likert1932technique} from 1 to 5, representing ``very unnatural'', ``somewhat unnatural'', ``neutral'', ``somewhat natural'', and ``very natural'', respectively. Note that the survey did not inform participants in advance about which video is attacked by \textbf{SLA}. In the semanticity test, videos were presented in pairs from the same source (one clean and one adversarial). Participants were informed which video was attacked and were asked to score solely based on whether the attack affects the video's semantics.
Similarly, they were required to rate the video from 1 to 5, representing ``very affected'', ``somewhat affected'', ``neutral'', ``somewhat unaffected'', and ``very unaffected'', respectively. Additionally, participants were paid \$0.8 and \$1.2 per question for the naturalness test and the semanticity test, respectively. The slightly higher payment for the latter compensated participants who were required to watch two videos to answer each question.

\begin{figure}[t]
   \centering
	\captionsetup{
		font={scriptsize}, 
	}
	\begin{adjustbox}{valign=t}
		\includegraphics[width=0.8\linewidth]{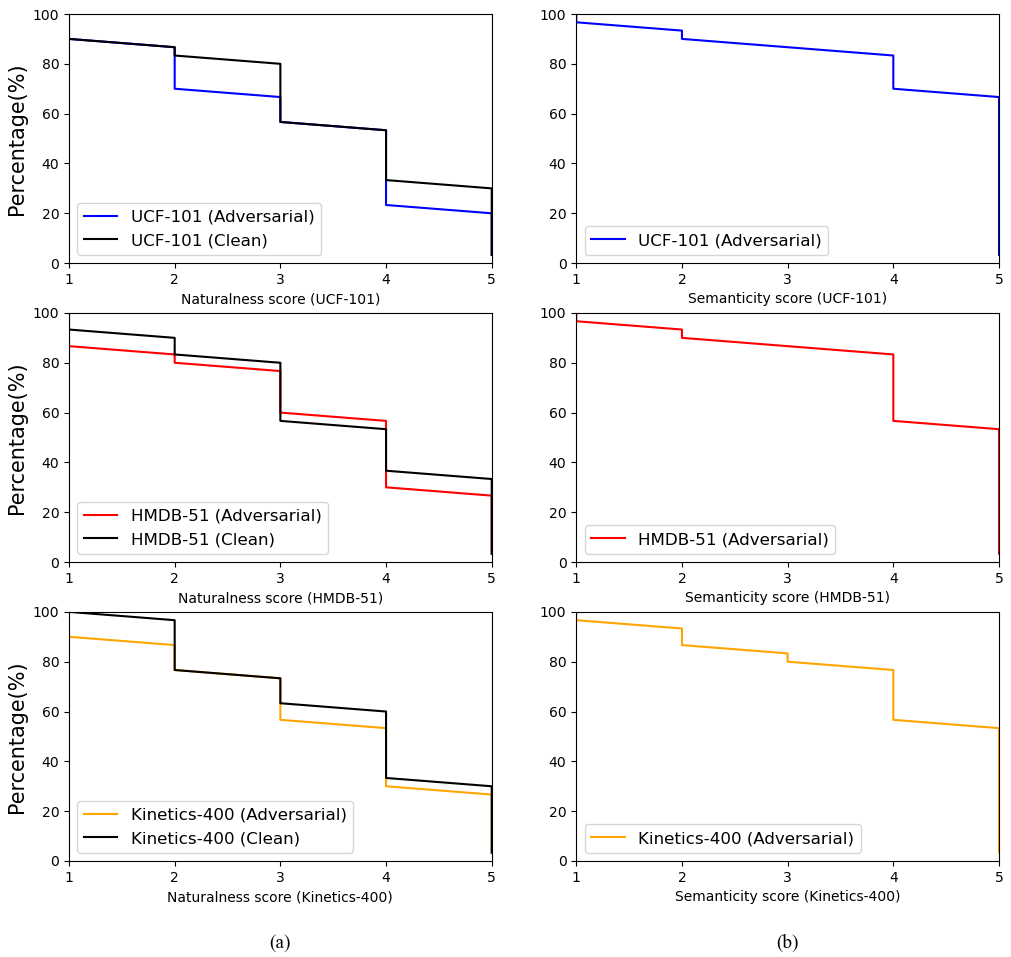}
	\end{adjustbox}
	\caption{Statistical distributions of naturalness and semanticity scores.}
	\label{figs:D_user}
    \vspace{-5mm}
\end{figure}

\subsubsection{Result Analysis}
\reffig{figs:D_user} shows the results for UCF-101, HMDB-51, and Kinetics-400. 
Each subplot illustrates the relationship between the score values and the proportion of videos rated above a given score relative to the total number of videos. Specifically, for a given score value in $\{1, 2, 3, 4, 5\}$, the highest corresponding proportion on the curve is taken as the intersection point. Thus, the closer the curve is distributed toward the upper-right region of the plot, the higher the overall evaluation of the videos by the participants.

\noindent\textbf{Naturalness.}
\reffig{figs:D_user}(a) shows the participants' naturalness ratings for adversarial and clean videos. 
Across all three datasets, the score distributions of the two types are closely aligned, with clean videos showing only a slight advantage. 
Notably, in HMDB-51, the proportion of adversarial videos rated at score 4 even exceeds that of clean videos. These results indicate that videos generated by \textbf{SLA} still appear natural and are difficult for humans to distinguish from clean ones. This can be largely attributed to \textbf{SLA}'s stylization strategy, which allows the added logos to blend harmoniously with the original video content rather than standing out conspicuously.

\noindent\textbf{Semanticity.}
\reffig{figs:D_user}(b) shows the results of the semantic impact of adversarial videos generated by \textbf{SLA}. Across all three datasets, more than 50\% of the adversarial videos were rated as ``very unaffected'', with this proportion exceeding 60\% on UCF-101.
Furthermore, fewer than 20\% of the adversarial videos in UCF-101 and HMDB-51 (30\% in Kinetics-400) exhibited semantic influence, suggesting that under RL guidance, \textbf{SLA} makes cautious decisions regarding logo size and positions, \eg keeping perturbations small and confined to video corners. Overall, the semanticity test validates the rationality of the designed reward function.

The two sets of user studies demonstrate the powerful capabilities of \textbf{SLA}-generated adversarial videos in terms of naturalness and semantics preservation.
Therefore, \textbf{SLA} simultaneously achieves strong attack effectiveness and good stealth, demonstrating practical value and warranting attention from the security community.

\section{DISCUSSION}
\label{sec:DISCUSSION}
\subsection{Ethical and Societal Impact}
Although \textbf{SLA} is an adversarial attack framework, its purpose is to reveal vulnerabilities in video AI systems and promote the development of more robust and trustworthy models. 
While \textbf{SLA} could theoretically be misused for malicious purposes such as bypassing content moderation, all experiments in this study were conducted under strict authorization and ethical supervision using only open-source models and public datasets.
We encourage the research community to view \textbf{SLA} as a diagnostic tool and to establish ethical guidelines and robustness benchmarks for adversarial research in video AI.

\subsection{Corner \& Center Validation}
To further validate that logos placed near the corners attract less human attention, we conduct an additional user study comparing their noticeability and perceived disturbance with those placed at the center. Following the same protocol described in Section~\ref{sec:User Study}, each participant was asked to view 60 randomly selected videos (20 with corner logos, 20 with center logos, and 20 clean). After viewing, participants had to rate two questions: (1) ``How easy is it to notice an anomalous element?'' on a scale from 1 (very easy) to 5 (very hard), and (2) ``To what extent does the anomalous element disturb your understanding?'' on a scale from 1 (not at all) to 5 (severely). Corner-logo videos achieve mean scores of 4.2 for noticeability and 2.1 for disturbance, whereas center-logo videos score 2.7 and 3.8, respectively, confirming that corner placement is significantly less noticeable and less disruptive.

\subsection{Dynamic Scenes}
Although a static logo may, in principle, become more noticeable when carried through a dynamic scene, we intentionally keep its position fixed to ensure implementation stability and maintain black-box query efficiency. 
From another perspective, the static logo can also serve as a copyright protection mechanism for the video. Therefore, we do not need to maintain any fixed spatial relationship between the logo and the object. Naturally, extending the static logo into a dynamically aware pasting strategy is a direction worth exploring in the future. For example, pasting a logo on a person's clothes and keeping it consistent with the moving figure by incorporating lightweight object detection or saliency analysis to dynamically adjust the logo position, would constitute a more stealthy perturbation.

\section{CONCLUSION}
\label{sec:CONCLUSION}
This paper presents a query-efficient adversarial attack against video classification models in service computing. In contrast to prior works that rely solely on style-transfer-based or patch-based perturbations, our study introduces a novel approach by integrating a stylized logo into the video's corner, acting as an adversarial patch. This method, dubbed \textbf{SLA}, has three key stages. First, a refined random-search-based strategy that combines SimBA with Square Attack to select the style reference is proposed to construct a style set. Next, an RL-driven style transfer is deployed to search for the logo's optimal attributes. Last, a perturbation optimizer named Logo-SSA optimizes the adversarial video in a step-by-step manner. Extensive experiments verify the superiority of \textbf{SLA} in both attack performance compared with five mainstream attacks and robustness against three existing defenses. \textbf{SLA}, as a novel method that combines style-transfer-based and patch-based approaches, can provide insights for stylized-patch attack research in the security community. Future work will explore potential defenses towards such stylized-patch attacks.

\section*{Acknowledgment}
The authors thank anonymous reviewers for their feedback that helped improve the paper. This work was supported in part by the Natural Science Foundation of Guangdong Province under Grant 2025A1515011946.

\small{
\bibliographystyle{IEEEtran}
\bibliography{reference}
}

\newpage
\appendix
\renewcommand{\thefigure}{S.\arabic{figure}}
\renewcommand{\thetable}{S.\arabic{table}}

\noindent\textbf{Hyperparameters Search.}
\textbf{SLA} has three important hyperparameters: style image number $N_s$, logo number $N_l$, and step size $\eta_p$ for optimizing perturbations in the third stage. Small-scale pre-experiments are conducted by randomly selecting 20 videos from UCF-101~\cite{soomro2012ucf101} to perform both targeted and untargeted attacks against C3D~\cite{tran2015learning}. A grid search is used to find the optimal parameter combination. 
The numerical results are analyzed from three aspects: $1)$ the impact of style image number $N_{s}$ on AQ$_1$, $2)$ the impact of logo number $N_{l}$ on AQ$_2$, and $3)$ the impact of perturbation step size $\eta_p$ on AQ$_3$. 

\reftable{tab:parameter_styleimg_num} shows that a larger $N_{s}$ directly causes more queries (AQ$_1$) in style reference selection (Stage 1). It expands the search space redundantly and consequently reduces attack efficiency. In contrast, a smaller $N_{s}$ provides limited selections for the following stage, which may result in a narrow search space for RL. Finally, we set $N_{s}$ as 5 in our experiment.

\reftable{tab:parameter_logo_num} reveals that $N_{l}$ tends to affect the convergence of queries (AQ$_2$) in RL-based logo style transfer (Stage 2). A moderate $N_{l}$ can help avoid search space redundancy and improve the efficiency of the optimization in the subsequent stage.
Finally, we set $N_{l}$ as 80 in our experiment.

\reftable{tab:parameter_step_size} indicates that $\eta_p$ is an important hyperparameter that significantly impacts the total number of queries in a single attack, since perturbation optimization (Stage 3) is the most query-consuming stage. A suitable $\eta_p$ can overcome the dilemma of local optima caused by pulling the samples back onto the $\ell_{\infty}$ ball. Finally, we set $\eta_p$ as 0.2 in our experiment.

\noindent\textbf{Quantitative Analysis on Video Quality.}
To evaluate the quality of the adversarial frames generated by \textbf{SLA}, we further calculate the Structural Similarity Index (SSIM) and the Fréchet Inception Distance (FID) between the original video and the adversarial video. A higher SSIM indicates greater structural similarity and visual consistency with the original video, while a lower FID reflects a closer distributional alignment in feature space, suggesting better perceptual quality of the adversarial video.

Tables~\ref{tab:ssim_fid_targeted} and~\ref{tab:ssim_fid_untargeted} present the numerical results. For \textbf{SLA}, SSIM remains between 0.80 and 0.86, 
indicating that pixel-level structures are largely preserved. The FID values range from 250 to 280 in most cases, remaining competitive with other methods and suggesting that the generated frames stay close to the real data distribution, which is considered an acceptable range for adversarial examples.

\begin{table}[h]
\centering
\caption{Results of \textbf{SLA} with various style image number $N_s$.}
\resizebox{1\linewidth}{!}{
\begin{tabular}{ccrrrrrrrrrrrrrrrr}
\toprule
\multirow{2}{*}[-0.5ex]{$N_s$} & \multicolumn{5}{c}{UCF-101-Targeted} & \multicolumn{5}{c}{UCF-101-Untargeted} \\
\cmidrule(r){2-6}\cmidrule(r){7-11}
& FR$\uparrow$ & AQ$_1$ & AQ$_2$ & AQ$_3$ & AQ$\downarrow$ & FR$\uparrow$ & AQ$_1$ & AQ$_2$ & AQ$_3$ & AQ$\downarrow$ \\
\midrule
1 & 40\% & 337.6 & 1,758.2 & 8,158.9 & 10,254.7 & 100\% & 1.0 & 613.1 & 1,270.5 & 1,884.6 \\
3 & 50\% & 1,474.3 & 1,800.4 & 7,114.7 & 10,389.4 & 100\% & 1.1 & 698.7 & 1,647.7 & 2,347.5 \\
5 & 60\% & 1,556.1 & 1,798.5 & 5,001.2 & 8,355.8 & 100\% & 1.2 & 560.6 & 1,127.4 & 1,689.2 \\
7 & 50\% & 1,646.2 & 1,615.8 & 6,891.4 & 10,153.4 & 100\% & 1.2 & 533.1 & 1,579.0 & 2,113.3 \\
9 & 60\% & 2,585.3 & 1,732.3 & 8,011.4 & 12,329.0 & 100\% & 1.4 & 590.7 & 1,624.7 & 2,216.8 \\
11 & 55\% & 3,158.1 & 1,795.2 & 9,044.5 & 13,997.8 & 100\% & 1.4 & 731.8 & 1,750.1 & 2,483.3 \\
\bottomrule
\end{tabular}
}

\label{tab:parameter_styleimg_num}
\end{table}

\begin{table}[h]  
\centering
\caption{Results of \textbf{SLA} with various logo number $N_l$.}
\resizebox{1\linewidth}{!}{
\begin{tabular}{ccrrrrrrrrrrrrrrrr}
\toprule
\multirow{2}{*}[-0.5ex]{$N_l$} & \multicolumn{5}{c}{UCF-101-Targeted} & \multicolumn{5}{c}{UCF-101-Untargeted} \\
\cmidrule(r){2-6}\cmidrule(r){7-11}
& FR$\uparrow$ & AQ$_1$ & AQ$_2$ & AQ$_3$ & AQ$\downarrow$ & FR$\uparrow$ & AQ$_1$ & AQ$_2$ & AQ$_3$ & AQ$\downarrow$ \\
\midrule
10 & 40\% & \multirow{6}{*}[-0ex]{1,899.8$\pm$382.7} & 1,662.1
 & 13,533.5 & 17,033.0 & 100\% & \multirow{6}{*}[-0ex]{1.2$\pm$0.2} & 563.6
 & 1,592.2 & 2,156.8 \\
20 & 55\% &  & 1,705.4 & 12,681.2 & 16,780.0 & 100\% &  & 617.4 & 1,451.3 & 2,069.7 \\
50 & 40\% &  & 1,683.0 & 17,290.5 & 21,102.1 & 100\% &  & 739.7 & 1,281.7 & 2,022.3 \\
80 & 60\% &  & 1,688.4 & 11,293.6 & 14,590.5 & 100\% &  & 511.3 & 663.7 & 1,175.9 \\
100 & 55\% &  & 1,702.3 & 14,039.6 & 16,995.6 & 100\% &  & 931.5 & 910.4 & 1,842.9 \\
120 & 55\% &  & 1,858.8 & 16,092.8 & 20,128.6 & 100\% &  & 634.9 & 1,335.9 & 1,971.7 \\
\bottomrule
\end{tabular}
}

\label{tab:parameter_logo_num}
\end{table}

\begin{table}[h]  
\centering
\caption{Results of \textbf{SLA} with various perturbation step size $\eta_p$.}
\resizebox{1\linewidth}{!}{
\begin{tabular}{ccrrrrrrrrrrrrrrrr}
\toprule
\multirow{2}{*}[-0.5ex]{$\eta_p$} & \multicolumn{5}{c}{UCF-101-Targeted} & \multicolumn{5}{c}{UCF-101-Untargeted} \\
\cmidrule(r){2-6}\cmidrule(r){7-11}
& FR$\uparrow$ & AQ$_1$ & AQ$_2$ & AQ$_3$ & AQ$\downarrow$ & FR$\uparrow$ & AQ$_1$ & AQ$_2$ & AQ$_3$ & AQ$\downarrow$ \\
\midrule
0.05 & 20\% & \multirow{6}{*}[-0ex]{2,049.8$\pm$199.3} & \multirow{6}{*}[-0ex]{1,686.3$\pm$93.8} & 10,528.7 & 13,112.6 & 100\% & \multirow{6}{*}[-0ex]{1.2$\pm$0.2} & \multirow{6}{*}[-0ex]{683.5$\pm$89.8} & 6,120.8 & 6,742.9 \\
0.1 & 35\% &  &  & 10,846.6 & 14,427.7 & 100\% &  &  & 4,866.8 & 5,747.4 \\
0.2 & 50\% &  &  & 4,632.7 & 8,393.9 & 100\% &  &  & 601.9 & 1,264.8 \\
0.3 & 40\% &  &  & 6,252.3 & 9,689.4 & 100\% &  &  & 1,207.3 & 1,823.2 \\
0.4 & 40\% &  &  & 9,297.9 & 13,432.8 & 100\% &  &  & 1,064.6 & 3,676.8 \\
0.5 & 40\% &  &  & 12,528.2 & 16,446.2 & 100\% &  &  & 1,516.3 & 2,179.7 \\
\bottomrule
\end{tabular}
}

\label{tab:parameter_step_size}
\end{table}

\begin{table*}[ht]
\centering
\caption{Quantitative evaluation on video quality of the targeted adversarial examples.}
\footnotesize
\resizebox{0.75\linewidth}{!}{
\begin{tabular}{cccccccccc}
\toprule
\multirow{2}{*}[-0.5ex]{Model} & 
\multirow{2}{*}[-0.5ex]{Attack} & \multicolumn{4}{c}{SSIM$\uparrow$} &  \multicolumn{4}{c}{FID$\downarrow$} \\
\cmidrule(r){3-6}\cmidrule(r){7-10}
& & UCF-101 & HMDB-51 & Kinetics-400 & Kinetics-700 & UCF-101 & HMDB-51 & Kinetics-400 & Kinetics-700 \\
\midrule
\multirow{6}{*}{C3D} 
& Sparse-RS \cite{croce2022sparse} & 0.7185 & 0.7313 & 0.7426 & 0.7612 & 269.64 & 288.61 & 266.64 & 241.88  \\
& Adv-watermark \cite{jia2020adv}  & 0.8153 & 0.8314 & 0.8566 & 0.8751 & 296.79 & 311.54 & 310.95 & 282.45 \\
& PatchAttack \cite{yang2020patchattack} & 0.7976 & 0.7523 & 0.7951 & 0.8134 & 279.59 & 304.21 & 297.79 & 275.91 \\
& BSC \cite{chen2022attacking} & 0.7923 & 0.8379 & 0.8494 & 0.8678 & 301.93 & 289.94 & 298.16 & 277.63 \\
& LogoStyleFool \cite{cao2024logostylefool} & 0.8056 & 0.8109 & 0.8289 & 0.8342 & 270.37 & 260.12 & 249.31 & 232.07  \\
& \textbf{SLA} & 0.8106 & 0.8199& 0.8361 & 0.8405 & 275.25 & 265.62 & 251.19& 212.62  \\
\midrule
\multirow{6}{*}{I3D} 
& Sparse-RS \cite{croce2022sparse} & 0.7185 & 0.7419 & 0.7364 & 0.7408 & 286.22 & 289.46 & 274.83 & 263.75  \\
& Adv-watermark \cite{jia2020adv}  & 0.8487 & 0.8285 & 0.8311 & 0.8356 & 310.42 & 301.64 & 295.21 & 284.33  \\
& PatchAttack \cite{yang2020patchattack} & 0.7685 & 0.7612 & 0.7899 & 0.7942 & 291.17 & 284.32 & 278.24 & 267.88 \\
& BSC \cite{chen2022attacking} & 0.7716 & 0.8021 & 0.8207 & 0.8251 & 307.18 & 291.18 & 310.62 & 299.45 \\
& LogoStyleFool \cite{cao2024logostylefool} & 0.8111 & 0.8012 & 0.8233 & 0.8287 & 272.32 & 244.17 & 261.12 & 252.44 \\
& \textbf{SLA} & 0.8273 & 0.8401& 0.8269 & 0.8315 & 263.33 & 257.62 & 270.31& 261.63\\
\bottomrule
\end{tabular}
}
\label{tab:ssim_fid_targeted}
\end{table*}

\begin{table*}[ht]
\centering
\caption{Quantitative evaluation on video quality of the untargeted adversarial examples.}
\footnotesize
\resizebox{0.75\linewidth}{!}{
\begin{tabular}{cccccccccccccccc}
\toprule
\multirow{2}{*}[-0.5ex]{Model} & 
\multirow{2}{*}[-0.5ex]{Attack} & \multicolumn{4}{c}{SSIM$\uparrow$} &  \multicolumn{4}{c}{FID$\downarrow$} \\
\cmidrule(r){3-6}\cmidrule(r){7-10}
& & UCF-101 & HMDB-51 & Kinetics-400 & Kinetics-700 & UCF-101 & HMDB-51 & Kinetics-400 & Kinetics-700 \\
\midrule
\multirow{6}{*}{C3D} 
& Sparse-RS \cite{croce2022sparse} & 0.7277 & 0.7406 & 0.7261 & 0.7455 & 278.74 & 282.92 & 284.17 & 254.33  \\
& Adv-watermark \cite{jia2020adv}  & 0.8257 & 0.8516 & 0.8439 & 0.8623 & 312.61 & 309.64 & 300.49 & 271.20 \\
& PatchAttack \cite{yang2020patchattack} & 0.7514 & 0.7862 & 0.7962 & 0.8157 & 303.64 & 296.41 & 292.41 & 269.47 \\
& BSC \cite{chen2022attacking} & 0.7844 & 0.8381 & 0.8036 & 0.8235 & 293.45 & 289.92 & 290.15 & 267.81 \\
& LogoStyleFool \cite{cao2024logostylefool} & 0.8086 & 0.8236 & 0.8361 & 0.8413 & 276.32 & 276.55 & 271.53 & 258.90  \\
& \textbf{SLA} & 0.8294 & 0.8329 & 0.8417 & 0.8461 & 266.51 & 255.92 & 264.52 & 252.38  \\
\midrule
\multirow{6}{*}{I3D} 
& Sparse-RS \cite{croce2022sparse} & 0.7213 & 0.7298 & 0.7495 & 0.7533 & 289.71 & 261.17 & 277.68 & 266.50  \\
& Adv-watermark \cite{jia2020adv}  & 0.8322 & 0.8354 & 0.8266 & 0.8307 & 294.76 & 281.51 & 278.62 & 268.91  \\
& PatchAttack \cite{yang2020patchattack} & 0.7624 & 0.7166 & 0.7588 & 0.7631 & 274.61 & 294.01 & 297.74 & 286.20 \\
& BSC \cite{chen2022attacking} & 0.7912 & 0.7709 & 0.8131 & 0.8178 & 303.29 & 291.16 & 283.54 & 297.63 \\
& LogoStyleFool \cite{cao2024logostylefool} & 0.8431 & 0.8332 & 0.8397 & 0.8442 & 262.61 & 262.94 & 259.46 & 250.81 \\
& \textbf{SLA} & 0.8306 & 0.8404 & 0.8217 & 0.8563 & 270.62 & 270.64 & 268.16 & 259.74 \\
\bottomrule
\end{tabular}
}
\label{tab:ssim_fid_untargeted}
\end{table*}

\begin{figure*}[ht]
\centering
	\captionsetup{
			font={scriptsize}, 
		}
	\centerline{\includegraphics[width=0.8\hsize]{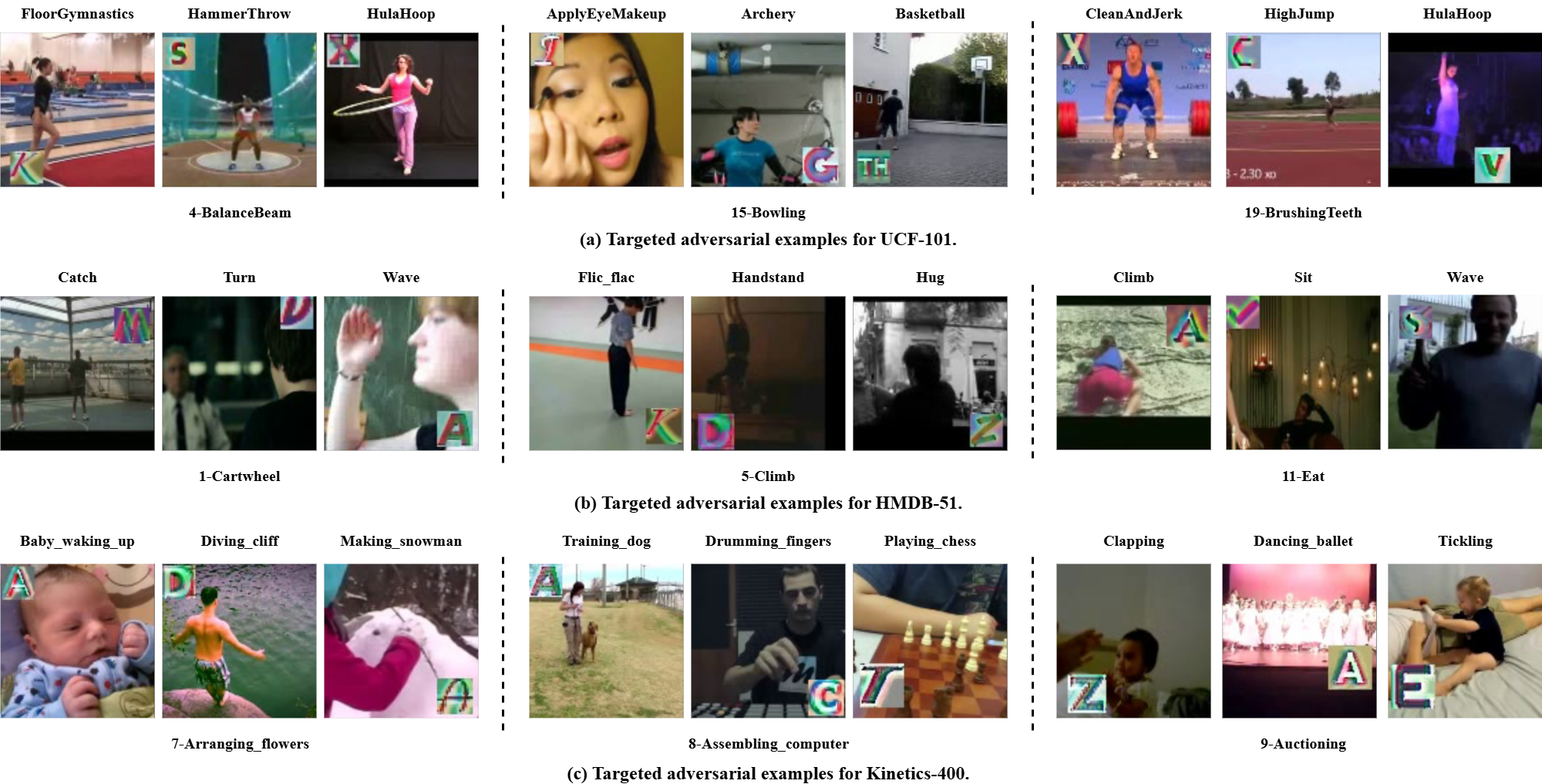}}
	\caption{Adversarial examples in targeted attacks. 
    The text above each image denotes the original video label, while the text below images denotes the target label for the adversarial video. 
    The number before the hyphen represents the label index. }
	\label{figs:Example_ours_T}
    \vspace{-3mm}
\end{figure*}

\begin{figure*}
\centering
	\captionsetup{
		font={scriptsize}, 
	}
		\includegraphics[width=0.6\linewidth]{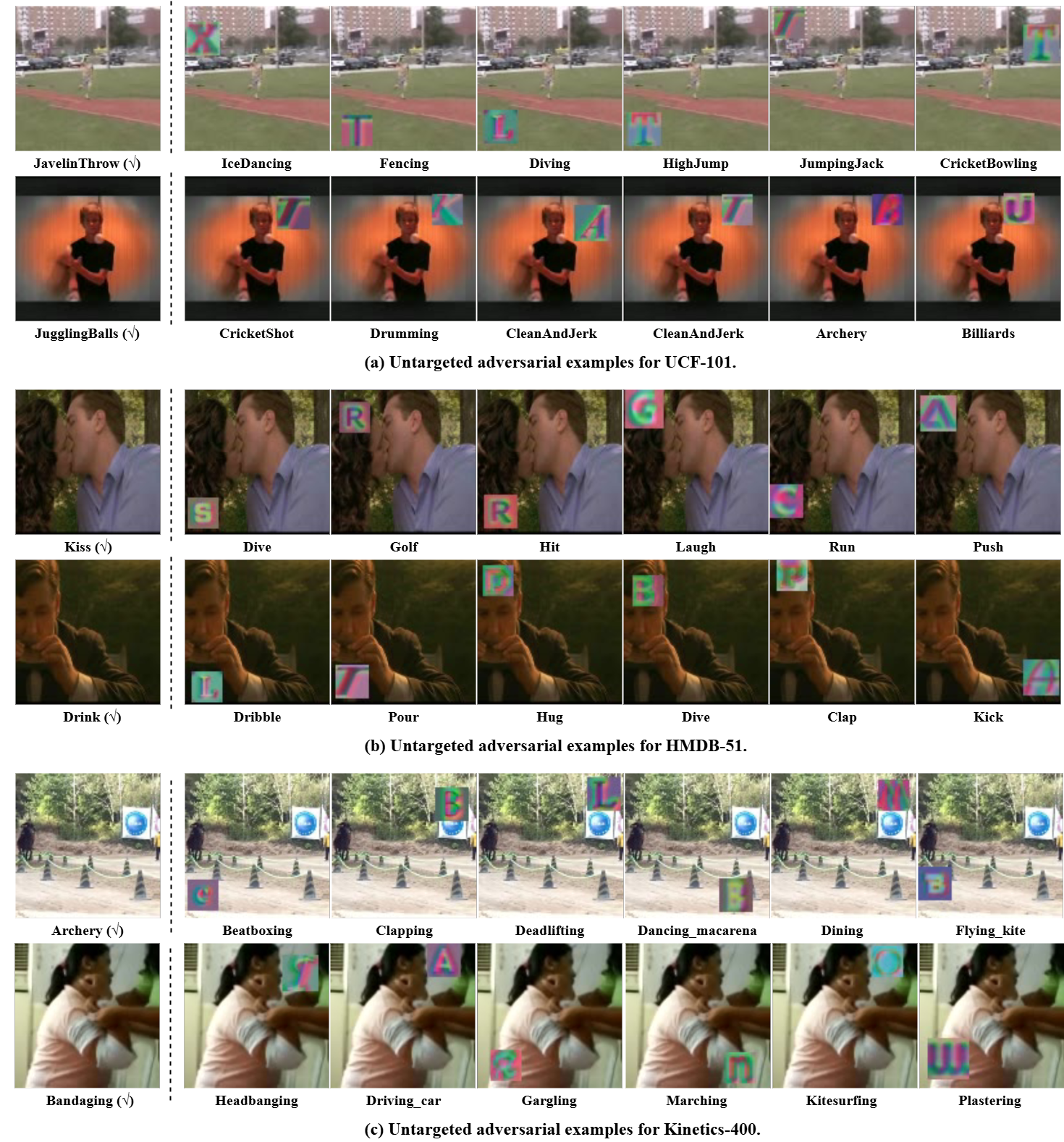}
	\caption{Adversarial examples in untargeted attacks. 
    The first column shows the original videos, with the texts below indicating their original label. The remaining six columns display the adversarial videos, where the text below denotes the misclassified label.
    }
	\label{figs:Example_ours_U}
    \vspace{-5mm}
\end{figure*}

\begin{figure*}[h]
	\captionsetup{
			font={scriptsize}, 
		}
	\centerline{\includegraphics[width=0.75\hsize]{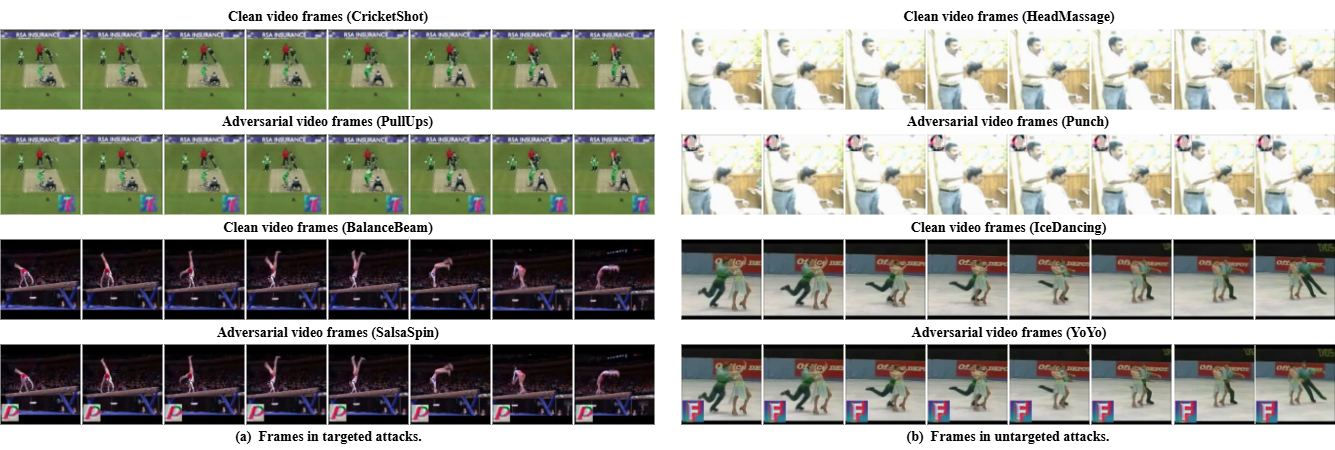}}
	\caption{Comparison of clean and adversarial video frames in targeted and untargeted attacks on UCF-101.}
	\label{figs:Frames_ucf101}
\end{figure*}

\begin{figure*}[h]
	\captionsetup{
			font={scriptsize}, 
		}
	\centerline{\includegraphics[width=0.75\hsize]{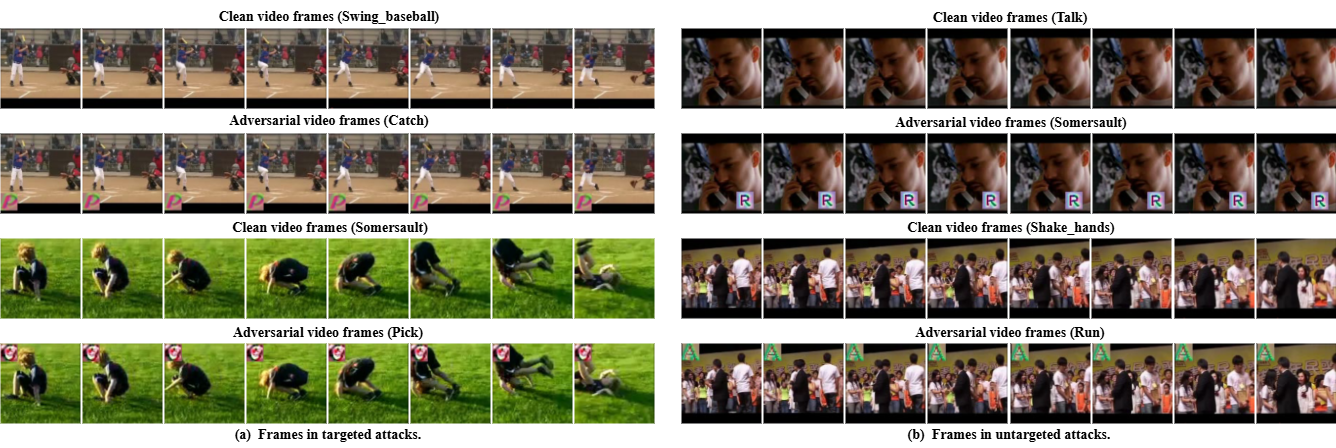}}
	\caption{Comparison of clean and adversarial video frames in targeted and untargeted attacks on HMDB-51.}
	\label{figs:Frames_hmdb51}
\end{figure*}

\begin{figure*}[h]
	\captionsetup{
			font={scriptsize}, 
		}
	\centerline{\includegraphics[width=0.75\hsize]{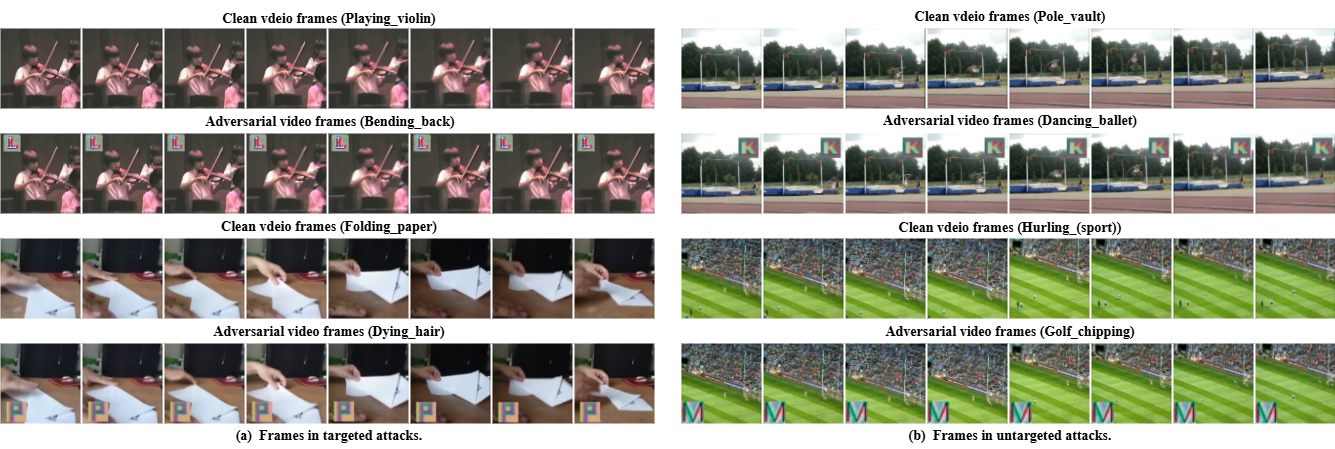}}
	\caption{Comparison of clean and adversarial video frames in targeted and untargeted attacks on Kinetics-400.}
	\label{figs:Frames_k400}
\end{figure*}

\begin{figure*}[h]
	\captionsetup{
			font={scriptsize}, 
		}
	\centerline{\includegraphics[width=0.75\hsize]{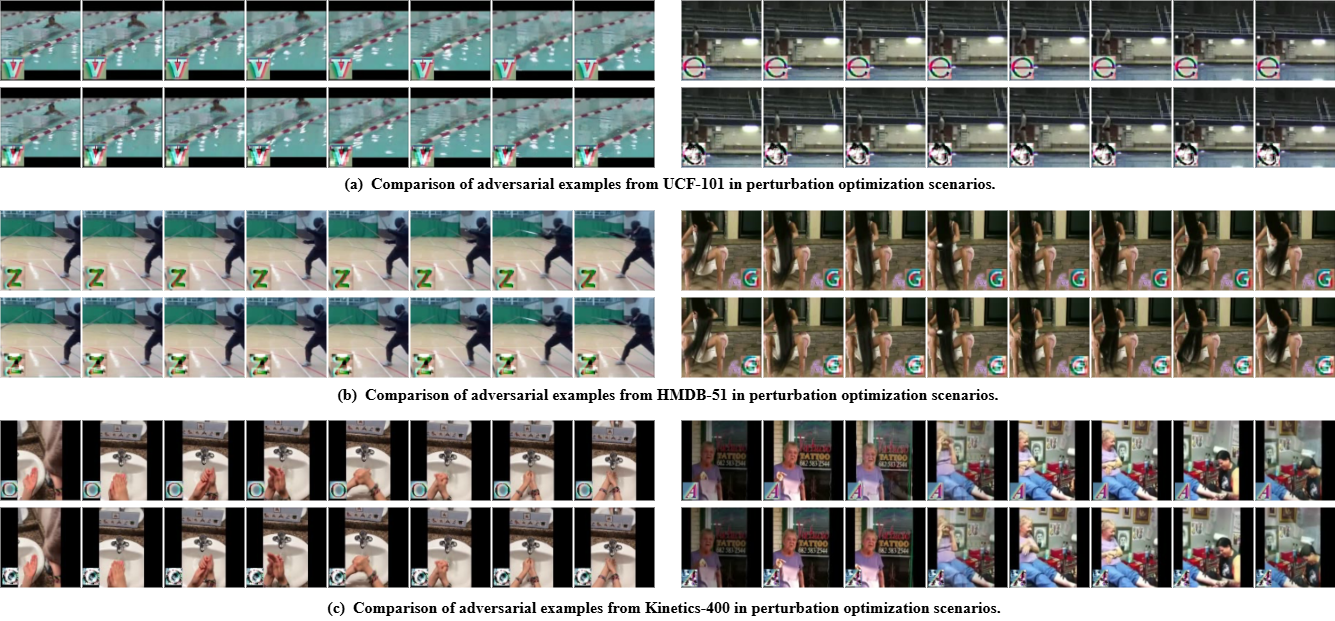}}
	\caption{Comparison of adversarial video frames when perturbation optimization is needed.}
	\label{figs:C_optimization}
\end{figure*}

\end{document}